\documentclass{article}

\usepackage{amsmath}
\usepackage{amsfonts}
\usepackage{microtype}
\usepackage{graphicx}
\usepackage{subcaption}
\usepackage[dvipsnames]{xcolor}
\usepackage{multirow}
\usepackage{booktabs} 
\usepackage[colorlinks=true,bookmarks=false,citecolor=blue,urlcolor=blue]{hyperref}

\usepackage{hyperref}

\usepackage[accepted]{icml2020}

\icmltitlerunning{Training Larger Networks for Deep RL}

\newcommand{\Tref}[1]{Table~\ref{#1}}

\newcommand{\fref}[1]{Fig.~\ref{#1}}
\newcommand{\Fref}[1]{Figure~\ref{#1}}
\newcommand{\sref}[1]{Sec.~\ref{#1}}

\newcommand{\Apref}[1]{Appendix~\ref{#1}}

\begin{document}
    \twocolumn[
        \icmltitle{Training Larger Networks for Deep Reinforcement Learning}
        \begin{icmlauthorlist}
            \icmlauthor{Kei Ota}{melco,tokyotech}
            \icmlauthor{Devesh K. Jha}{merl}
            \icmlauthor{Asako Kanezaki}{tokyotech}
        \end{icmlauthorlist}
        \icmlaffiliation{melco}{Mitsubishi Electric, Kanagawa, Japan}
        \icmlaffiliation{merl}{Mitsubishi Electric Research Labs, Cambridge, USA}
        \icmlaffiliation{tokyotech}{Tokyo Institute of Technology, Tokyo, Japan}
        \icmlcorrespondingauthor{Kei Ota}{Ota.Kei@ds.MitsubishiElectric.co.jp}
        \icmlkeywords{Reinforcement Learning}
        \vskip 0.3in
    ]

    \printAffiliationsAndNotice{}  

    \begin{abstract}
The success of deep learning in the computer vision and natural language processing communities can be attributed to training of very deep neural networks with millions or billions of parameters which can then be trained with massive amounts of data. However, similar trend has largely eluded training of deep reinforcement learning (RL) algorithms where larger networks do not lead to performance improvement. Previous work has shown that this is mostly due to instability during training of deep RL agents when using larger networks. In this paper, we make an attempt to understand and address training of larger networks for deep RL. We first show that naively increasing network capacity does not improve performance. Then, we propose a novel method that consists of 1) wider networks with DenseNet connection, 2) decoupling representation learning from training of RL, 3) a distributed training method to mitigate overfitting problems. Using this three-fold technique, we show that we can train very large networks that result in significant performance gains. We present several ablation studies to demonstrate the efficacy of the proposed method and some intuitive understanding of the reasons for performance gain. We show that our proposed method outperforms other baseline algorithms on several challenging locomotion tasks.


\end{abstract}
    \section{Introduction}\label{sec:intro}
We have witnessed huge improvements in the fields of computer vision (CV) and natural language processing (NLP) in the last decade~\cite{krizhevsky2012imagenet,he2016deep,huang2017densely,devlin2019bert,brown2020language}. These developments could largely be attributed to training of very large neural networks with millions (or even billions or trillions) of parameters which can be trained using massive amounts of data and an appropriate optimization technique to stabilize training. In general, the motivation of training larger networks comes from the intuition that larger networks allow better solutions as they increase the search space of possible solutions. Having said that, neural network training largely relies on the ability to find good minimizers of highly non-convex loss functions. These loss functions are also governed by the choices of network architecture, batch size, etc. This has also driven a lot of research in these communities towards understanding the underlying reasoning for performance gains~\cite{lu2017expressive,zhang2017understanding,nguyen2017loss,li2018visualizing}.


\begin{figure}[t]
	\begin{minipage}[]{0.5\columnwidth}
		\centering
		\includegraphics[height=4.4truecm]{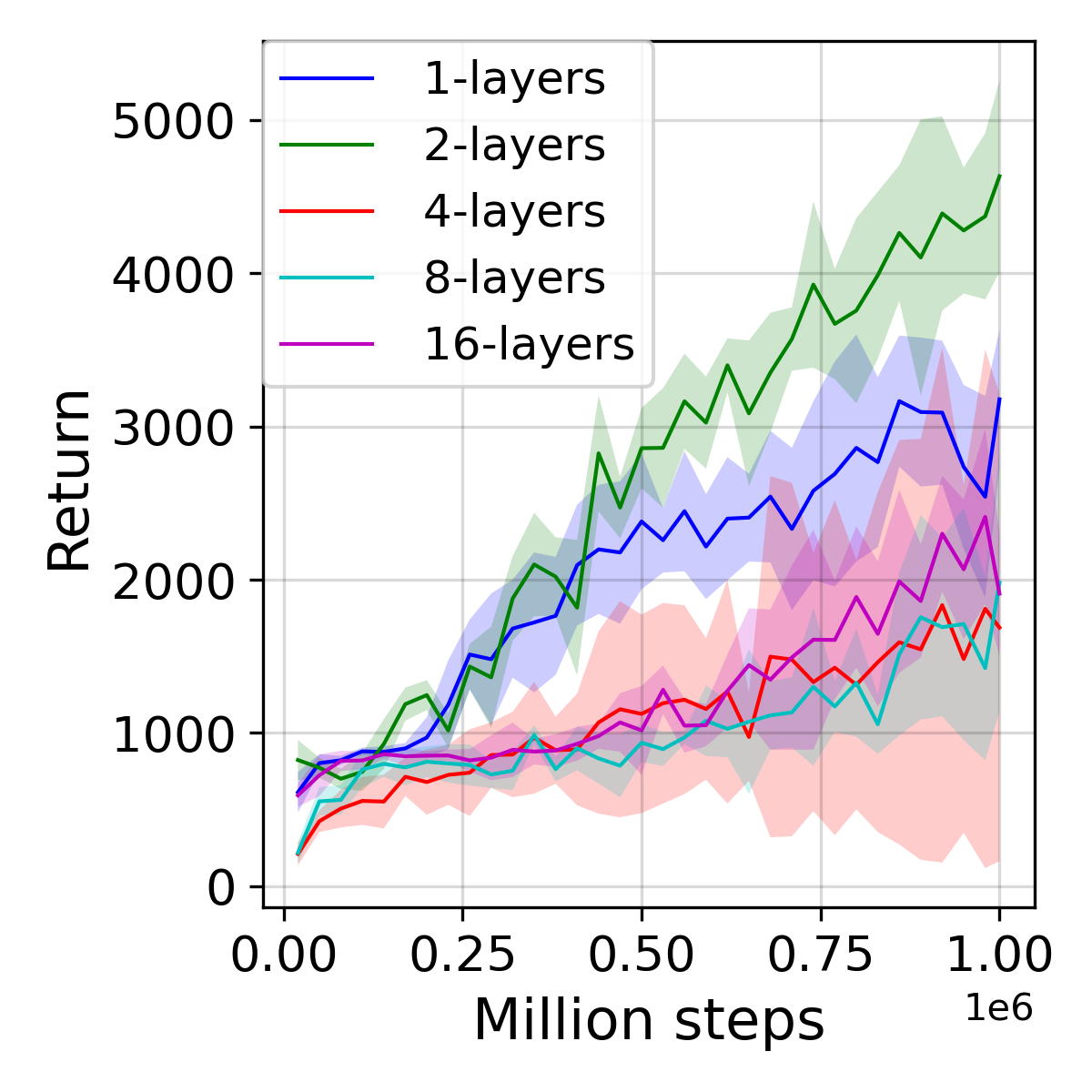}
		\vskip -0.05in
		\subcaption{Average return.}\label{fig:sac_deeper_return}
	\end{minipage}
	\begin{minipage}[]{0.48\columnwidth}
		\centering
		\includegraphics[height=4.4truecm]{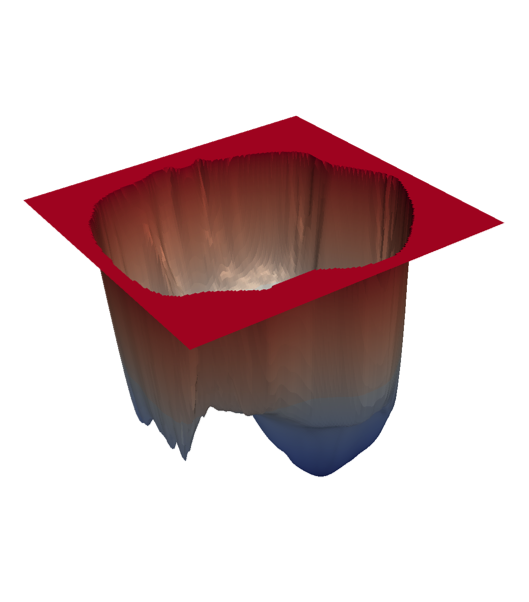}
		\vskip -0.05in
		\subcaption{Loss surface.}\label{fig:sac_deeper_curvature}
	\end{minipage}
	\vskip -0.05in
    \caption{Training curves of SAC agents with different number of layers on Ant-v2 environment, and the loss function surface of the deepest (16-layers) Q-network. The training curves suggest that simply building a deeper MLP with fixed number of units (256) does not improve the performance of DRL while building a larger network is generally effective in supervised learning. Motivated by this, we conduct an extensive study on how to train larger networks that contribute for performance gain for RL agents.}
	\label{fig:sac_deeper}
\end{figure}

In a striking contrast, Deep Reinforcement Learning (DRL) community has not reported similar trend with regards to training larger networks for RL. It has been reported in some studies that deep RL agents experience instability while training with larger networks~\cite{henderson2018deep,hasselt2018deep,achiam2019towards,sinha2020d2rl}. 
As an example, in \fref{fig:sac_deeper}, we show the results of an Soft Actor Critic (SAC)~\cite{haarnoja2018soft} agent that uses Multi-layered Perceptron (MLP) for function approximation with increasing number of layers while fixing its unit size to 256 (also notice the loss surface).
These plots show that using deeper networks naively leads to poor performance for a deep RL agent. Consequently, using larger networks for training deep RL networks is not fully understood, and thus is limiting in several ways. As a result, most of the reported work in literature end up using similar hyperparameters (i.e., network structure, number and size of layers, etc.). Our work is motivated by this limitation, and we make an attempt to explore the interplay between the size, structure, training and performance of deep RL agents to provide some intuition and guidelines for using larger networks.

In light of these facts, we present a large-scale study and provide empirical evidence for using larger networks for training DRL agents.
We first highlight the challenges that one might come across while using larger networks for training deep RL agents.
To circumvent these problems, we integrate a three-fold approach: decoupling feature representation from RL to efficiently produces high dimensional features, employing DenseNet architecture to propagate richer information, and using distributed training methods to collect more on-policy transitions to reduce overfitting.
Our method is a novel architecture that combines these three elements, and we demonstrate our proposed method significantly improves the performance of RL agents in continuous control tasks.
We also conduct an ablation study to show what component contributes the performance gain.
Our contributions can be summarized as follows:
\begin{itemize}
    \item We conduct a large scale study on employing larger networks for DRL agents, and empirically show that, in contrary to deeper networks, wider networks can improve performance.
    \item We propose a novel network architecture that synergistically combines recently proposed techniques to stabilize the training: decoupling representation learning from RL, DenseNet architecture, and distributed training, to demonstrate it significantly improves performance.
    \item We analyze the performance gain of our method using metrics of effective ranks of features as well as visualization of loss function landscape of RL agents.
\end{itemize}


    \begin{figure*}[th]
        \centering
        \includegraphics[width=0.8\textwidth]{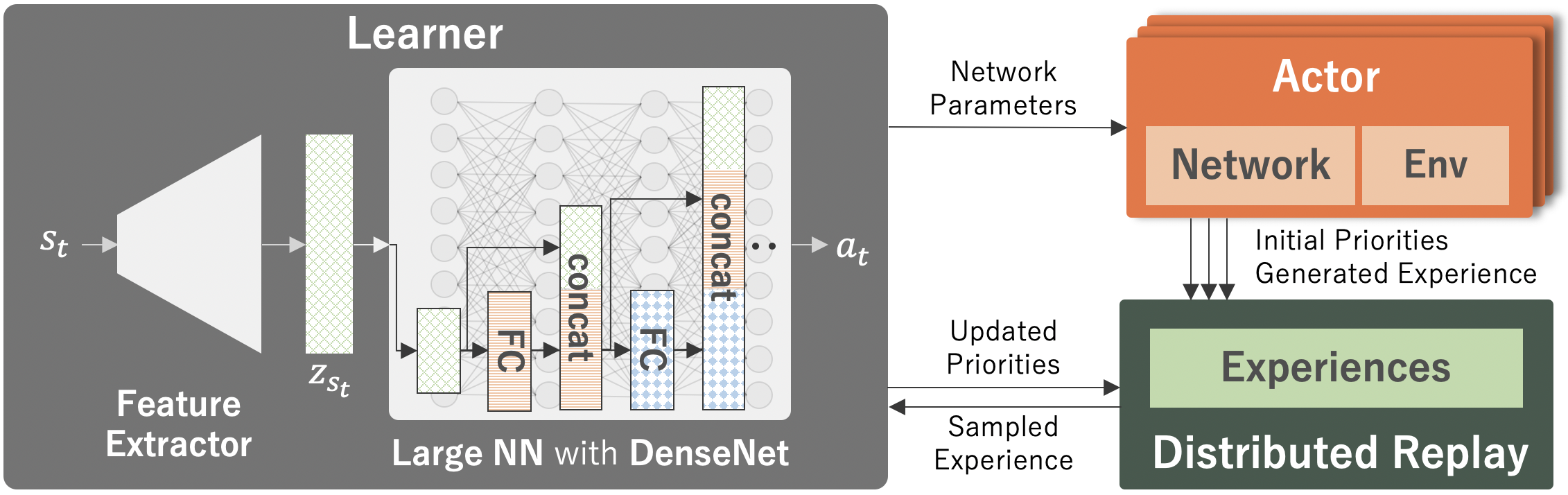}
        \caption{Proposed architecture to train larger networks for deep RL agents. 
        We combine three elements. Firstly, we decouple representation learning from RL to extract an informative feature $z_{s_t}$ from the current state $s_t$ using a feature extractor network that is trained using an auxiliary task of predicting the next state $s_{t+1}$.
        Secondly, we use large networks using DenseNet architecture, which allows stronger feature propagation. Finally, we employ the Ape-X-like distributed training framework to mitigate the overfitting problems which tends to happen in larger networks, and enables to collect more on-policy data that can improve performance.
        FC refers to a fully-connected layer.}
        \label{fig:architecture}
    \end{figure*}

    \section{Related Work}
Our work is broadly motivated by \cite{henderson2018deep} that empirically demonstrates DRL algorithms are vulnerable to different training choices like architectures, hyperparameters, activation functions, etc. The paper compares performance on different number of units and layers, and demonstrates larger networks do not consistently improve performance.
This is contrary to our intuition considering the recent progress on solving computer vision tasks such as ImageNet~\cite{deng2009imagenet}: larger and more complex network architectures have proven to achieve better performance~\cite{krizhevsky2012imagenet,he2016deep,huang2017densely,tan2019efficientnet}.

\citet{sutton2018reinforce} identifies a \emph{deadly triad} of \emph{function approximation}, \emph{bootstrapping}, and \emph{off-policy learning}. When these three properties are combined, learning can be unstable, and potentially diverge with the value estimates becoming unbounded.
Some prior works have challenged to mitigate this problem, including target networks~\cite{mnih2015human}, double Q-learning~\cite{van2016deep}, n-step learning~\cite{hassel2018rainbow}, etc.
Our challenge of training larger networks is specifically related to function approximation, however, as the deadly triad is entangled in a complex manner, we also have to deal with the other problems.
As for the network size, some studies investigate the effect of making network larger for continuous control task using MLP~\cite{fu2019diagnosing,achiam2019towards} and Atari games using CNN~\cite{hasselt2018deep}, and concluded the larger networks tend to perform better, but also become unstable and prone to diverge more.
\citet{andrychowicz2020matters} and \citet{liu2021regularization} performed similar study on on-policy methods and showed too small or large networks can cause significant drop in performance of the policy.
While these studies are limited to relatively small size (hundreds of units with several layers), we will have more thorough study on much larger networks, combination of state representation learning, and employing different network architectures.

To build a large network, unsupervised learning has been used to learn powerful representations for downstream tasks in natural language processing~\cite{devlin2019bert,radford2019language} and computer vision~\cite{he2020momentum,chen2020a}.
In the context of RL, auxiliary tasks such as predicting the next state conditioned on the past state(s) and action(s) have been widely studied to improve the sample efficiency of the RL algorithms~\cite{jaderberg2017reinforcement,shelhamer2017loss,ha2018worldmodels}.
For the state-input setting, researchers have generally focused on learning a good representation that produces low dimensional features~\cite{munk2016learning,lesort2018state}.
Contrary to that, \citet{ota2020can} proposes the use of online feature extractor network (OFENet) that intentionally increases input dimensionality, and demonstrates that larger feature size enables to improve RL performance on both sample efficiency and control performance.
We leverage this idea and use larger input (or feature) for RL agents as well as using larger networks for the policy and the value function networks.
    \section{Method}\label{sec:methods}
While recent studies suggest that larger networks for DRL agents have potential to improve performance, it is non-trivial to alleviate some potential issues that lead to instability when using larger networks to train RL agents.

Our method is based on two main key ideas: allowing better feature propagation using good network architectures and using huge amounts of more on-policy data using distributed training to avoid overfitting in larger networks.
We first obtain good features apart from RL using an auxiliary task, and then propagate the features more efficiently by employing the DenseNet~\cite{huang2017densely} architecture. Also, we use a distributed RL framework that can mitigate the potential overfitting problem.
In the following, we describe in detail the three elements we use for training larger networks for deep RL agents. Our proposed approach is shown as a schematic in~\fref{fig:architecture}.



\subsection{Decoupling Representation Learning from RL} \label{sec:method_ofenet}
While the simplicity of learning whole networks in an end-to-end fashion is appealing, updating all parameters of a large network using just a scalar reward signal can result in very inefficient training~\cite{stooke2020decoupling}.
Decoupling unsupervised pretraining from downstream tasks is common in computer vision~\cite{he2020momentum,henaff2020data}.
Taking inspiration from this, we adopt the online feature extractor network (OFENet)~\cite{ota2020can} to learn meaningful features separately from training of RL.

OFENet learns representation vectors of states $z_{s_t}$ and state-action pairs $z_{s_t,a_t}$, and provides them to the agent instead of original inputs $s_t$ and $a_t$, which gives significant performance improvements on continuous robot control tasks.
As the representation vectors $z_{s_t}$ and $z_{s_t,a_t}$ are designed to have much higher dimensionality than original inputs, OFENet matches our philosophy of providing larger solution space that allows us to find better policy. 
The representations can be obtained by learning the mappings $z_{s_t}=\phi_s(s_t)$ and $z_{s_t,a_t}=\phi_{s,a}(s_t,a_t)$, which have parameters $\theta_{\phi_s}, \theta_{\phi_{s,a}}$ by using an auxiliary task of predicting the next state $s_{t+1}$ from the current state and action representation $z_{s_t,a_t}$ as:
\begin{equation}\label{eqn:aux_loss}
    L_\text{aux} = \mathbb{E}_{(s_t, a_t) \sim p,\pi} \left[ \|f_\text{pred}(z_{s_t,a_t}) - s_{t+1} \|^2 \right],
\end{equation}
where $f_\text{pred}$ is represented as a linear combination of the representation $z_{s_t, a_t}$. The learning of the auxiliary task is done concurrently with the learning of the downstream RL task.
In our experiments, we allow input dimensionality much bigger than previously presented in~\cite{ota2020can}.
Furthermore, we also increase the network size of RL agents (see \ref{appendix:architectures} for the number of network parameters used in our experiments).
For more details, interested readers are referred to~\cite{ota2020can}.

\subsection{Distributed Training} \label{sec:method_dist_sample}
In general, larger networks need more data to improve accuracy of function approximation~\cite{deng2009imagenet,hernandez2021scaling} and mitigate overfitting problem~\cite{bishop2006pattern}.
MLP with a large number of hidden layers is in particular known to cause an over-fitting to training data, which often results in inferior performance to shallow networks~\cite{bengio2007greedy,ramchoun2017new}.
In the context of RL, while we are training and evaluating on the same environment, there is still problem of overfitting: the agent is only trained on limited trajectories it has experienced, which cannot cover the whole state-action space of the environment~\cite{liu2021regularization}.
\citet{fu2019diagnosing} showed overfitting to the experience replay exists, and~\citet{fedus2020revisiting} empirically showed having more on-policy data in replay buffer, i.e. collecting more than one transition while updating policy one time can improve performance of RL agent.

In light of these studies, we employ distributed RL framework, which leverages distributed training architectures that decouples learning from collecting transitions by utilizing many actors running in parallel on separate environment instances~\cite{horgan2018distributed,kapturowski2018recurrent}.
In particular, we use Ape-X~\cite{horgan2018distributed} framework, where a single learner receives experiences from distributed prioritized replay~\cite{schaul2016prioritized}, and multiple actors collect transitions in parallel (see \fref{fig:architecture}).
This helps increase the number of data that are close to the current policy, i.e. more on-policy data, which can improve performance of off-policy RL agents~\cite{fedus2020revisiting} and mitigate rank collapse issues of Q-networks~\cite{kumar2021implicit}.
One difference is that we do not use the RL algorithm used in~\cite{horgan2018distributed}, but instead use standard off-policy RL algorithms: SAC~\cite{haarnoja2018soft} and Twin Delayed Deep Deterministic policy
gradient algorithm (TD3)~\cite{fujimoto2018addressing} in our experiments.

\subsection{Network Architectures}\label{sec:method_architecture}
Tremendous developments have been made in the computer vision community in designing sophisticated architectures that enable training of very large networks~\cite{he2016deep,huang2017densely,tan2019efficientnet}. 
\citet{huang2017densely} proposed Dense Convolutional Network (DenseNet) that has a skip connection that directly connects each layer to all subsequent layers as: $y_{i} = f_i^\text{dense}([ y_0, y_1, ..., y_{i-1}])$, where $y_i$ is the output of the $i^\text{th}$ layer, thus all the inputs are concatenated into a single tensor. Here, $f_i^\text{dense}$ is a composite function which consists of a sequence of convolutions, Batch Normalization (BN)~\cite{ioffe2015batch}, and an activation function.
An advantage of DenseNet is its improved flow of information and gradients throughout the network, which makes the large networks easier to train.
We borrow this architecture to train large networks for RL agents.

Although using DenseNet architecture for DRL agents is existing, it has not been fully explored yet. D2RL~\cite{sinha2020d2rl} employs a modified DenseNet architecture which concatenates the state or the state-action pair to each hidden layer of the MLP networks except the last linear layer.
Contrary to this modified version, \citet{ota2020can} exactly follows the original DenseNet: it uses the dense connection that concatenates all the outputs of the previous layer instead of only the state or the state-action pair for training only OFENet, and is not used for training RL agents.
We also follow the original DenseNet architecture as done in~\cite{ota2020can} to represent the policy and the value function networks.
The schematic of the DenseNet architecture is also shown in \fref{fig:architecture}. 
We omit BN for SAC agent because we found it inhibits improving performance.
\section{Experiments}
In this section, we present results of numerical experiments in order to answer some relevant underlying questions posed in this paper. In particular, we answer the following questions.

\begin{itemize}
    \item Can RL agents benefit from usage of larger networks during training? More concretely, can using larger networks lead to better policies for DRL agents? 
    \item What characterizes a \textit{good} architecture which facilitates better performance when using larger networks?
    \item Can our method work across different RL algorithms as well as different tasks?
\end{itemize}

\paragraph{Experimental settings}
We run all each experiment independently with five seeds, and the average and $\pm 1$ standard deviation results will be reported, which are solid lines and shaded regions when we show training curves.
The horizontal axis of a training curve is the number of gradient steps, which is not identical to the number of steps an agent interacts with an environment only when we use the distributed replay. 
The network architectures, optimizers, and hyperparameters are the same as used in their original papers~\cite{haarnoja2018soft,fujimoto2018addressing,ota2020can} unless otherwise noted.
We used single NVIDIA Tesla V100 GPU with Xeon Gold 6148 Processor.
\Apref{appendix:exp_details} shows more details of experimental settings.

\paragraph{Evaluation metrics}
We evaluate the experimental results on two metrics: average return and the recently proposed \emph{effective ranks}~\cite{kumar2021implicit} of the features matrices of Q-networks.
\citet{kumar2021implicit} showed that MLPs used for approximating policy and value functions that use bootstrapping leads to reduction in the effective rank of the feature, and this rank collapse for the feature matrix results in poorer performance.
We will show the effective rank of the features in the penultimate layer of the Q-networks to evaluate whether our proposed architecture can alleviate the rank collapse issue.

\subsection{Does increasing the size of networks fail to improve performance?}\label{sec:exp_depth_width}
In the first set of experiments, we try to investigate if increasing network size always leads to poor performance. We quantitatively measure the effectiveness of increasing the network size by changing the number of units $N^\text{unit}$ and layers $N^\text{layer}$ while the other parameters are fixed.

\begin{figure}[t]
	\begin{minipage}[]{0.5\columnwidth}
		\centering
		\includegraphics[height=4.4truecm]{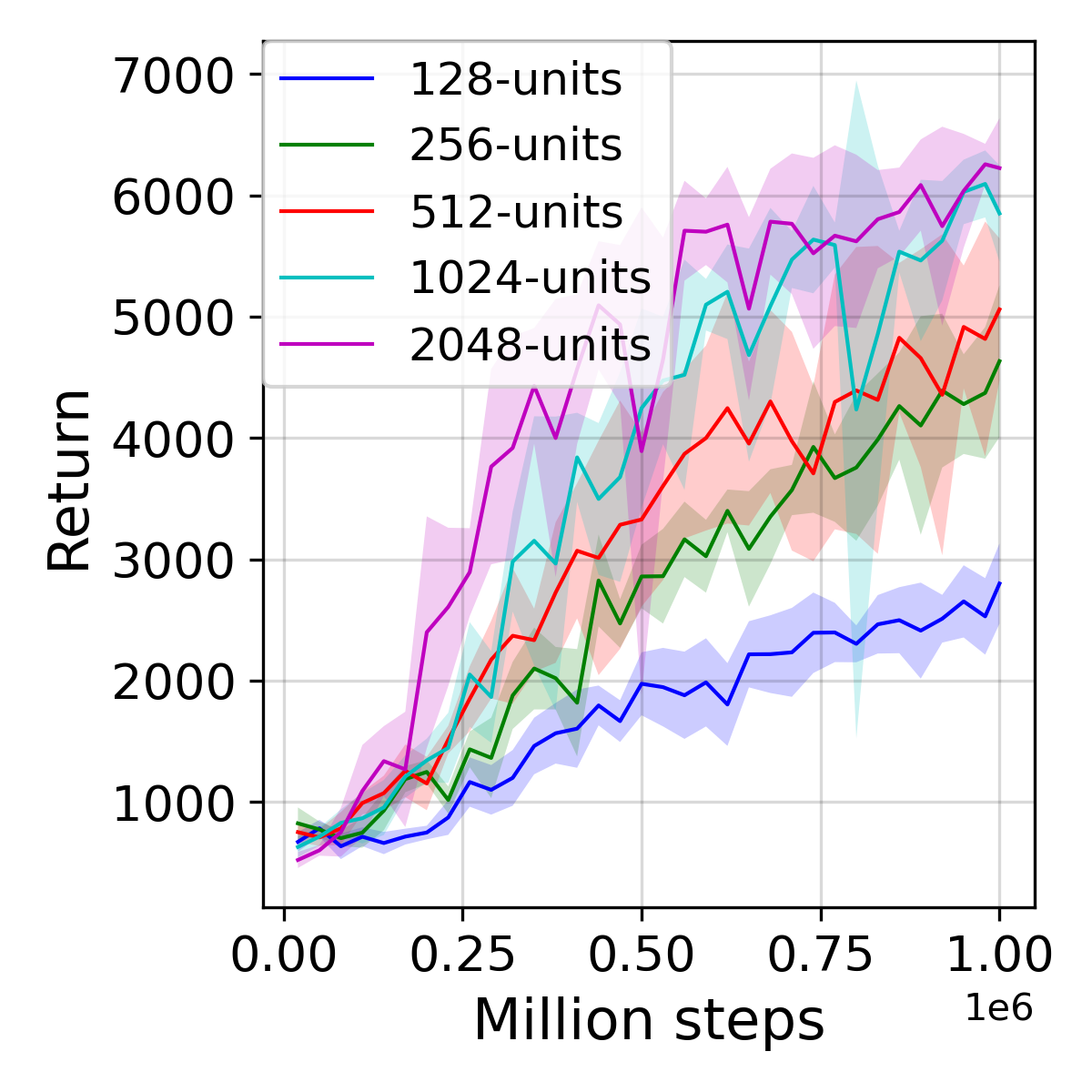}
		\vskip -0.05in
		\subcaption{Average return.}\label{fig:sac_wider_return}
	\end{minipage}
	\begin{minipage}[]{0.48\columnwidth}
		\centering
		\includegraphics[height=4.4truecm]{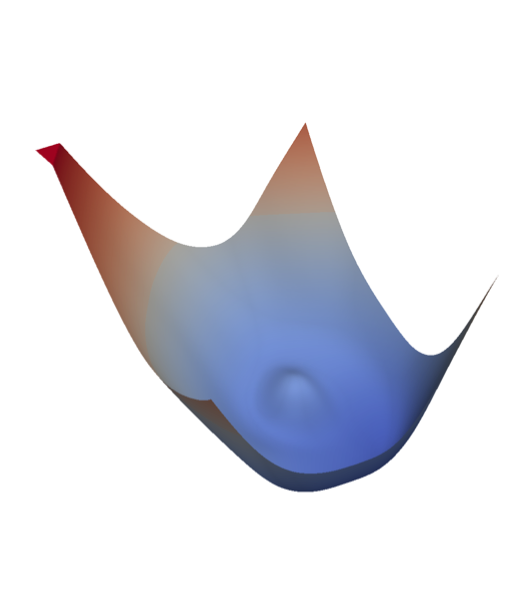}
		\vskip -0.05in
		\subcaption{Loss surface.}\label{fig:sac_wider_curvature}
	\end{minipage}
	\vskip -0.05in
    \caption{Training curves of the SAC agent with different number of units on Ant-v2 environment and the loss function surface of the widest (2048-units) Q-network. This shows the performance consistently improves when using wider MLPs.}
	\label{fig:sac_wider}
    \vskip 0.05in
    \centering
    \includegraphics[width=0.95\columnwidth]{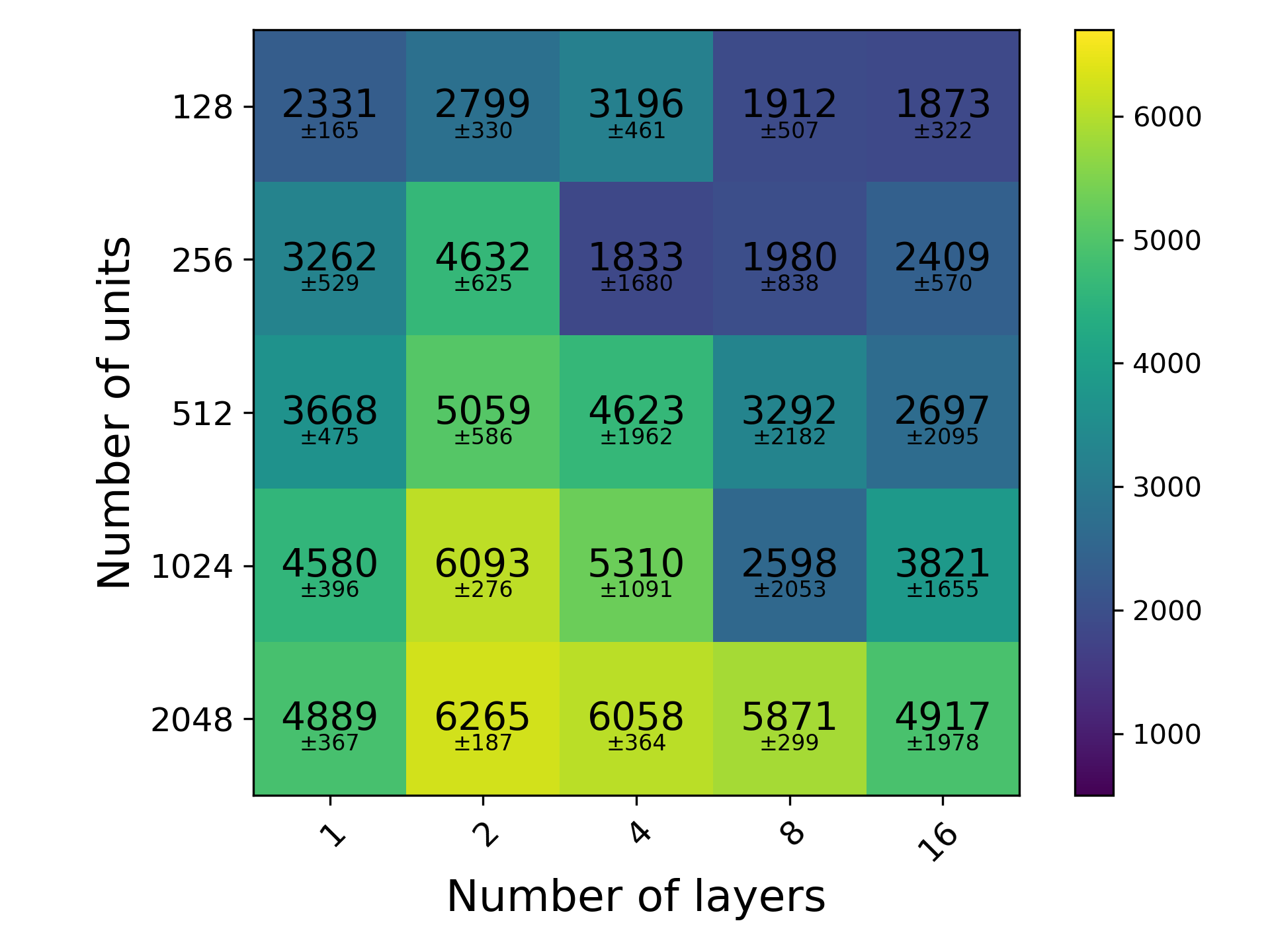}
    \vskip -0.15in
    \caption{Grid search results of maximum average return at one-million training steps over different number of units and layers for SAC agent on Ant-v2 environment. This demonstrates a deeper MLP (see horizontally) does not consistently improve performance while a wider MLP (see vertically) generally does.}
    \label{fig:grid_search_ant_sac_woofe_raw}
\end{figure}

\Fref{fig:sac_deeper_return} shows the training curves when increasing the number of layers while the unit size is fixed to $N^\text{unit}=256$. As we described in~\sref{sec:intro}, we observe that the performance becomes worse as the network becomes deeper.
In \fref{fig:sac_wider_return}, we show the effect of increasing the number of units while the number of layers is fixed to $N^\text{layer}=2$.
Contrary to the results when making the network deeper, we can observe consistent improvement when making the network wider.
In order to investigate more thoroughly, we also conduct a grid search, where we sample each parameter of the network from $N^\text{unit}\in \left\{ 128, 256, 512, 1024, 2048 \right\}$, and $N^\text{layer} \in \left\{ 1, 2, 4, 8, 16 \right\}$ and evaluate the performance in \fref{fig:grid_search_ant_sac_woofe_raw}.
We can see the monotonic improvement in performance when widening networks on almost all depth of the network.

This result is in line with the general belief that training deeper networks is, in general, harder and is more susceptible to choice of hyperparameters~\cite{bengio2007greedy,ramchoun2017new}.
This could be attributed to vanishing gradient problem with increasing number of layers~\cite{bengio1994learning}. However, we found that the reason that deeper networks are harder to train than wider network cannot be attributed to vanishing gradient, rather it results from the sharpness of the loss surface curvatures~\cite{li2018visualizing}.
We show the loss surface of the deeper network ($N^\text{layer}=16, N^\text{unit}=256$) in \fref{fig:sac_deeper_curvature} and the wider network ($N^\text{layer}=2, N^\text{unit}=2048$) in \fref{fig:sac_wider_curvature} by using the visualization method proposed in~\cite{li2018visualizing} with the loss of TD error of Q-functions of SAC agents (see \Apref{appendix:deeper_wider_surface} for more details).
These figures show that wider networks have nearly convex surface while deeper networks have more complex loss surface which could be susceptible to choice of hyperparameters~\cite{li2018visualizing}. Comparison of deeper and wider networks have also been done in 
  several works~\cite{wu2019wider,nguyen2017loss,li2018visualizing}, where wider networks are prone to have more generalization capability due to their smooth loss functions.

From these results, we observe and conclude that larger networks can be effective in improving deep RL performance. In particular, we achieve consistent performance gains when widening individual layers instead of going deeper. Consequently,
 we fix the number of layers to $N^\text{layer}=2$, and only change the number of units to learn larger networks in the following experiments.

\subsection{Architecture Comparison}\label{sec:architecture_comparison}
In the next set of experiments, we try to investigate the role of synergistic combination of connectivity architecture, state-representation and distributed training in allowing usage of larger networks for training deep RL agents. A brief introduction to these techniques is described in \sref{sec:methods}.

\paragraph{Connectivity architecture}
We first compare four connectivity architectures: standard MLP, MLP-ResNet, MLP-DenseNet, 
and MLP-D2RL, which is a recently proposed architecture to improve RL performance.
MLP-ResNet is a modified version of Residual Networks (ResNet)~\cite{he2016deep,he2016identity}, which has a skip-connection that bypasses the non-linear transformations with an identity function: $y_{i} = f_i^\text{res}(y_{i-1}) + y_{i-1}$, where $y_i$ is the output of the $i^\text{th}$ layer, and $f_i^\text{res}$ is a residual module, which consists of fully connected layer and nonlinear activation function.
An advantage of this architecture is that the gradient can flow directly through the identity mapping from top layers to bottom layers.
MLP-D2RL is identical to~\cite{sinha2020d2rl}, and MLP-DenseNet is our proposed architecture that is defined in~\sref{sec:method_architecture}.
We compare these four architectures on both small network ($N^\text{unit}=128$, denoted by \emph{S}) and large networks ($N^\text{unit}=2048$, denoted by \emph{L}).

\begin{figure}[t]
	\begin{minipage}[]{0.623\columnwidth}
		\centering
		\includegraphics[height=5.6truecm]{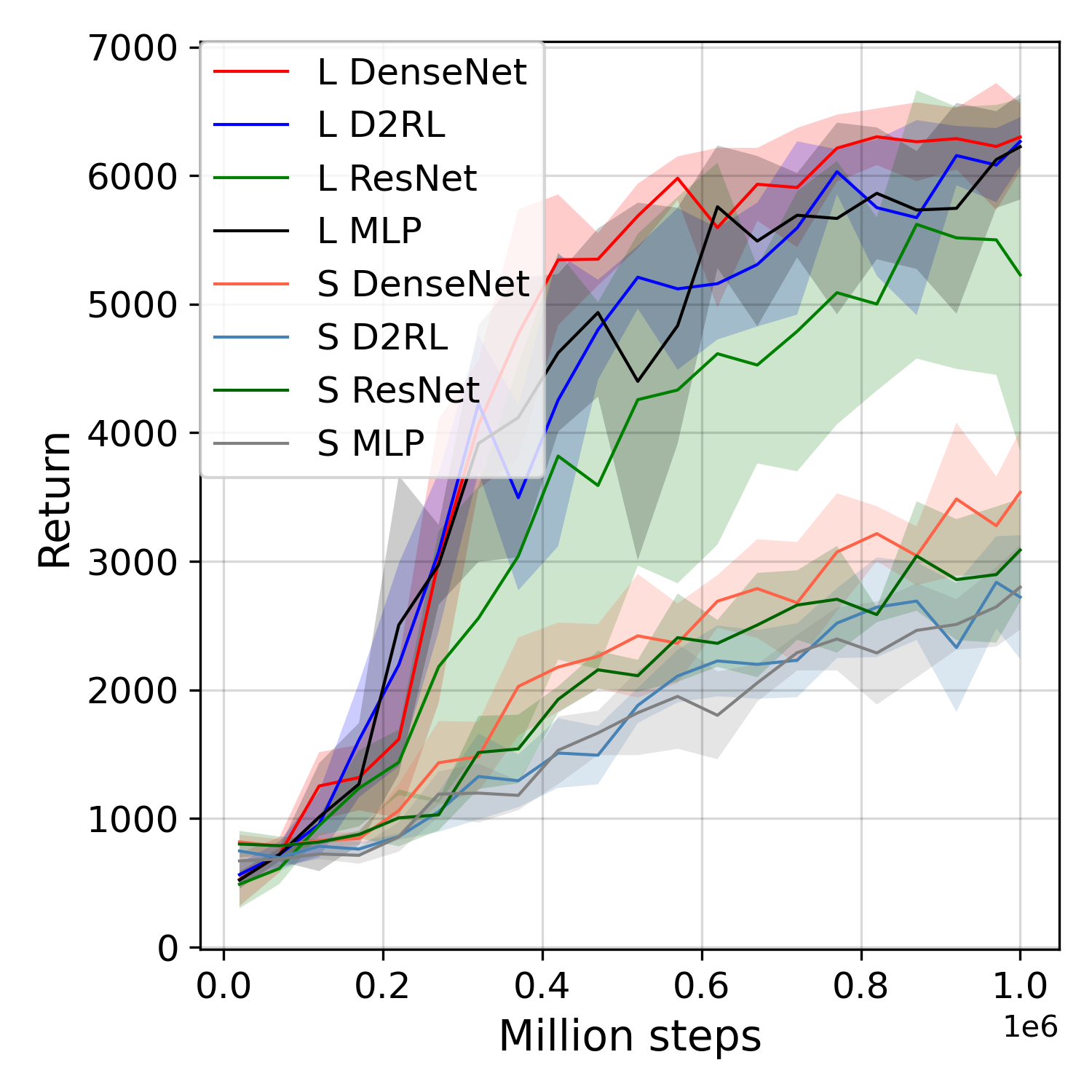}
		\vskip -0.05in
		\subcaption{Average return.}\label{fig:connectivity_comparison_return}
	\end{minipage}
	\begin{minipage}[]{0.36\columnwidth}
		\centering
		\includegraphics[height=5.6truecm]{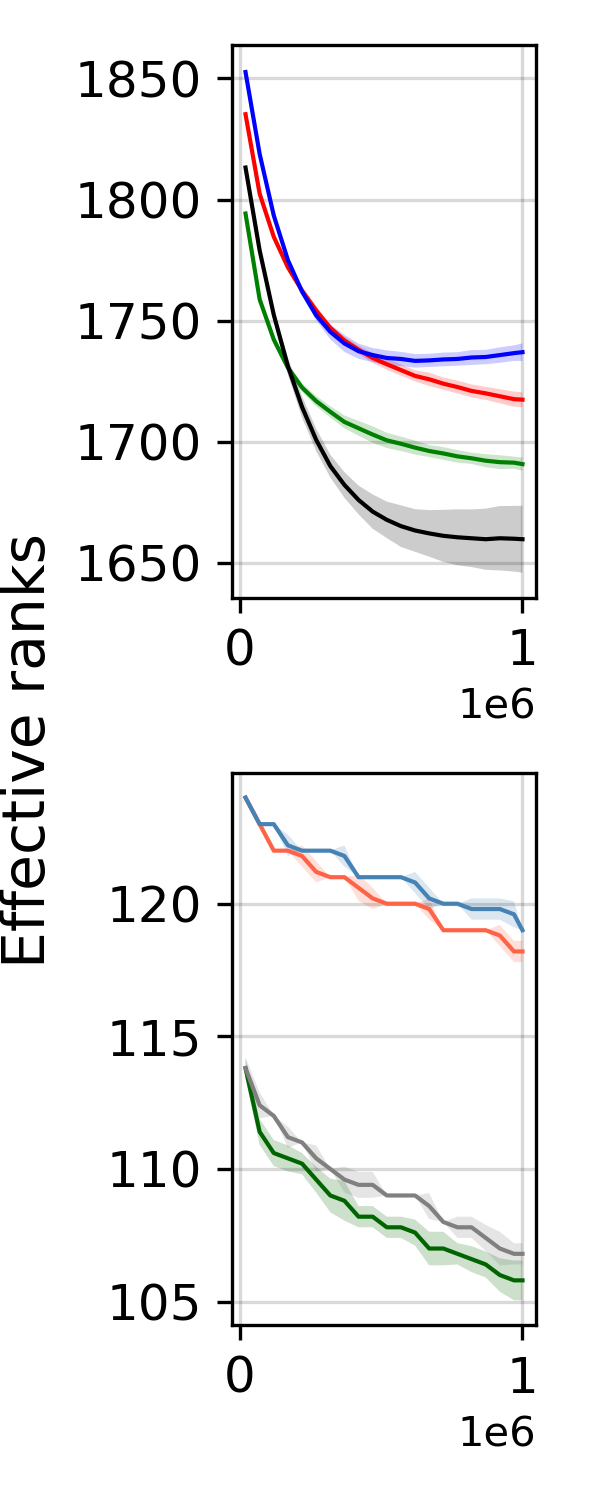}
		\vskip -0.05in
		\subcaption{Effective ranks.}\label{fig:connectivity_comparison_rank}
	\end{minipage}
	\vskip -0.05in
    \caption{Comparison of connectivity architecture on Ant-v2. Our proposed DenseNet architecture produces the best return on both large ($N^\text{unit}=2048$, denoted by \emph{L}) and small ($N^\text{unit}=128$, \emph{S}) networks while mitigating rank collapse as good as MLP-D2RL.}
	\label{fig:connectivity_comparison}
\end{figure}

\Fref{fig:connectivity_comparison} shows the training curves of average return in \fref{fig:connectivity_comparison_return}, and the effective ranks in \fref{fig:connectivity_comparison_rank}.
The results show that our MLP-DenseNet achieves the highest return on both small and large networks, while mitigating rank collapse comparable to MLP-D2RL.
This shows that the MLP-DenseNet is the best architecture among these four choices, and thus we employ this architecture for both the policy and the value function network in the following experiments.

\paragraph{Decoupling representation learning from RL}
Next, we evaluate the effectiveness of using OFENet (see \sref{sec:method_ofenet})  to decouple representation learning from RL.
In order to evaluate the performance on different network size, we sample the number of units from $N^\text{units} \in \left \{ 256, 1024, 2048 \right\}$, which we respectively denote \emph{S}, \emph{M}, and \emph{L}, and 
compare these against the baseline SAC agents, which do not use OFENet and are trained only from a scalar reward signal.
In other words, the baseline agents are identical to the DenseNet architecture of the previous connectivity comparison experiment.

\begin{figure}
	\begin{minipage}[]{0.623\columnwidth}
		\centering
		\includegraphics[height=5.6truecm]{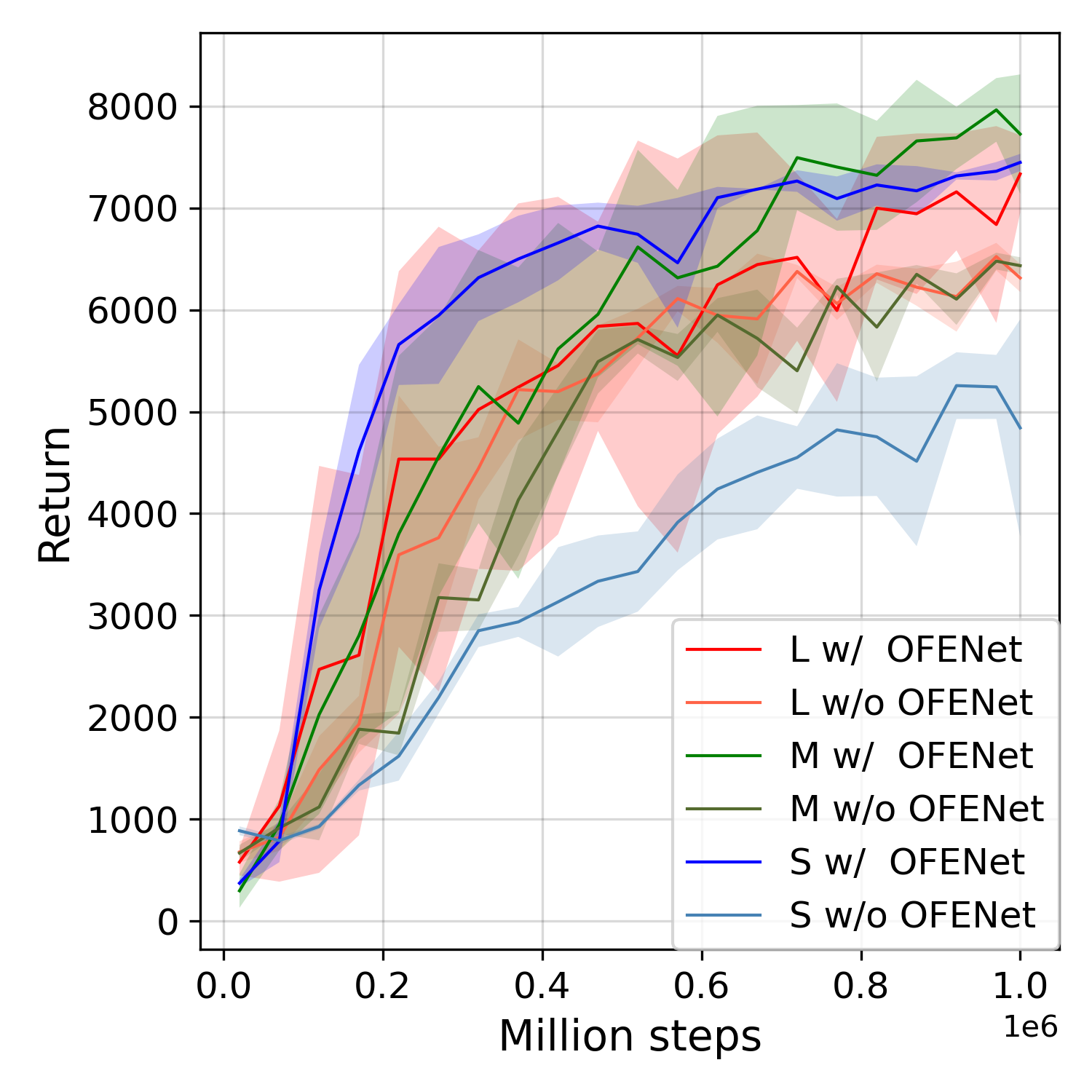}
		\vskip -0.05in
		\subcaption{Average return.}\label{fig:ofe_comparison_return}
	\end{minipage}
	\begin{minipage}[]{0.36\columnwidth}
		\centering
		\includegraphics[height=5.6truecm]{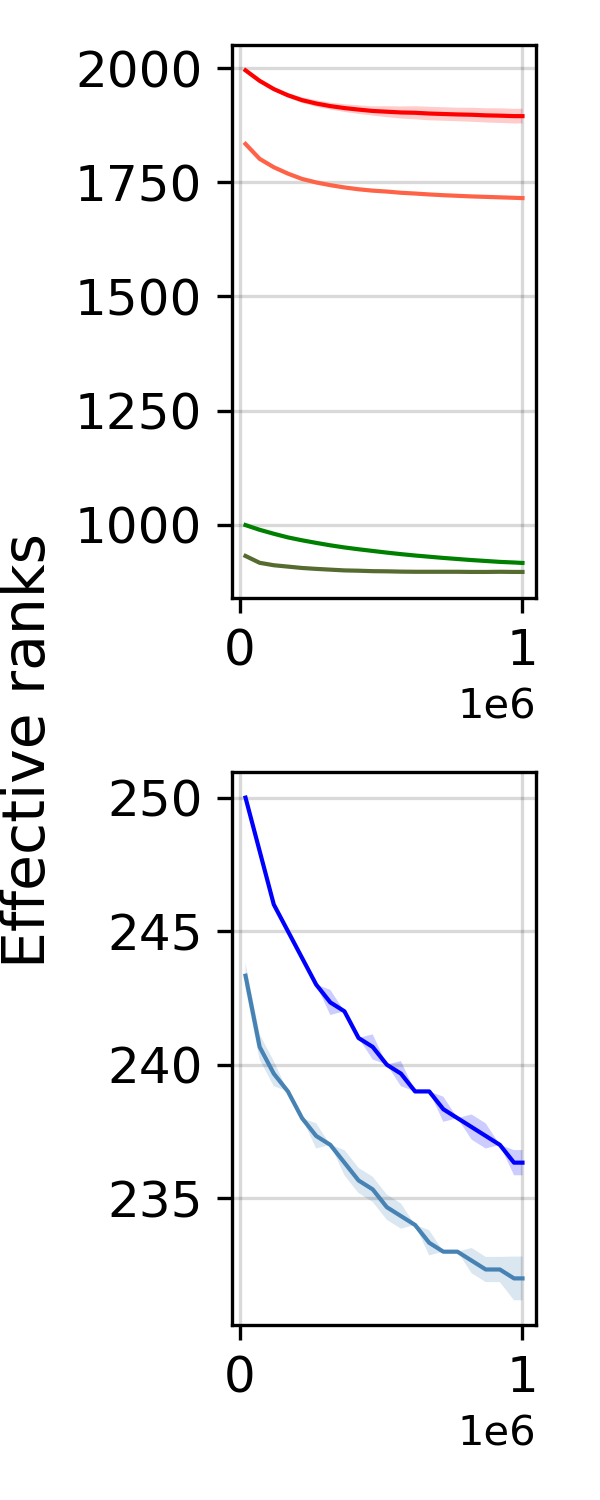}
		\vskip -0.05in
		\subcaption{Effective ranks.}\label{fig:ofe_comparison_rank}
	\end{minipage}
	\vskip -0.05in
    \caption{Training curves of w/ and w/o OFENet on Ant-v2. This shows decoupling representation learning from RL is generally effective across different size of the networks in terms of both control performance and mitigating rank collapse issues.}
	\label{fig:ofenet}
    \centering
    \includegraphics[width=0.95\columnwidth]{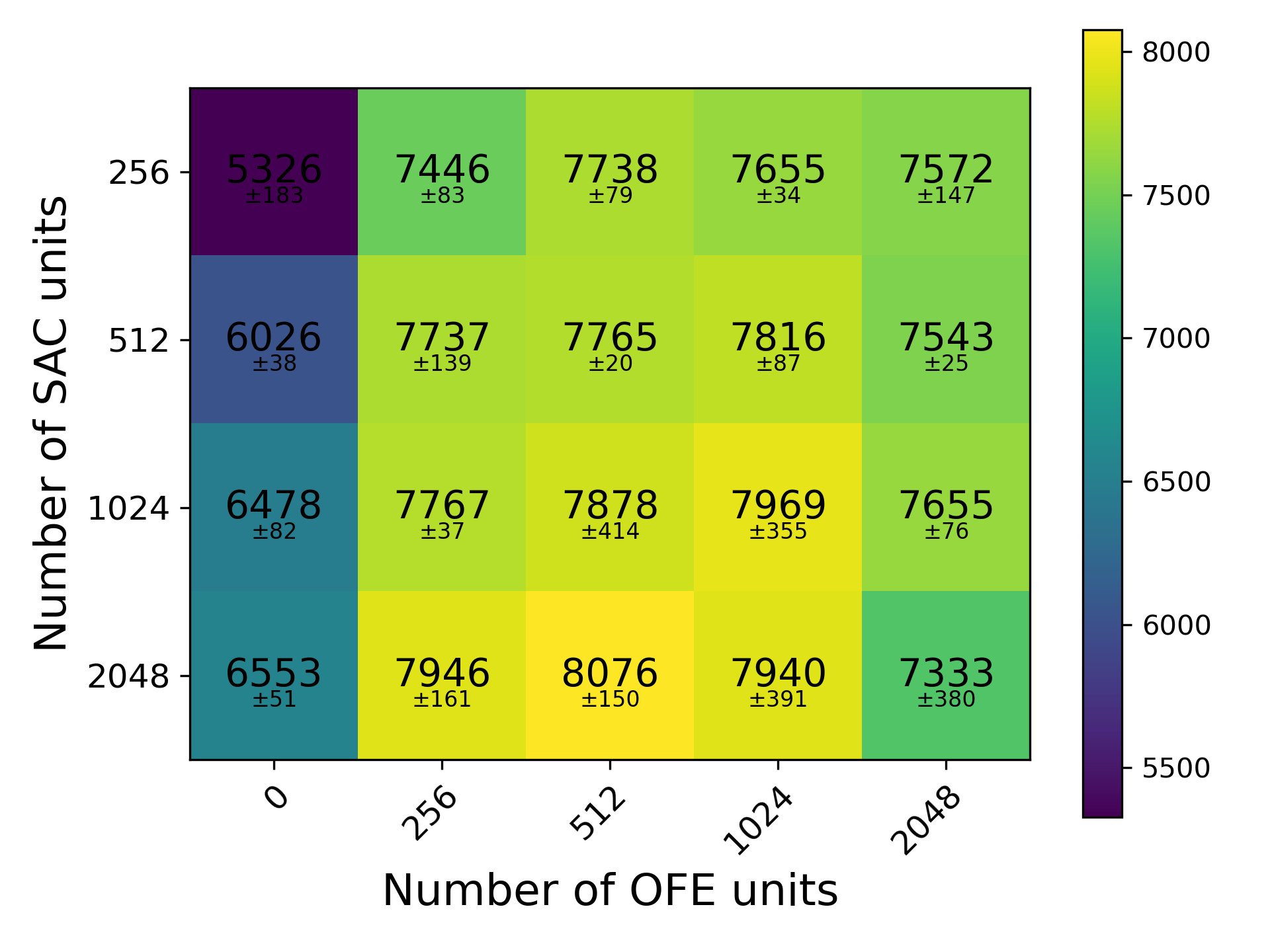}
    \vskip -0.2in
    \caption{Grid search results of average maximum return over different number of units between SAC and OFENet. OFENet can improve performance on almost all settings, but saturates around the return of 8000.}
    \label{fig:ofe_sac_grid_search}
    \includegraphics[width=0.95\columnwidth]{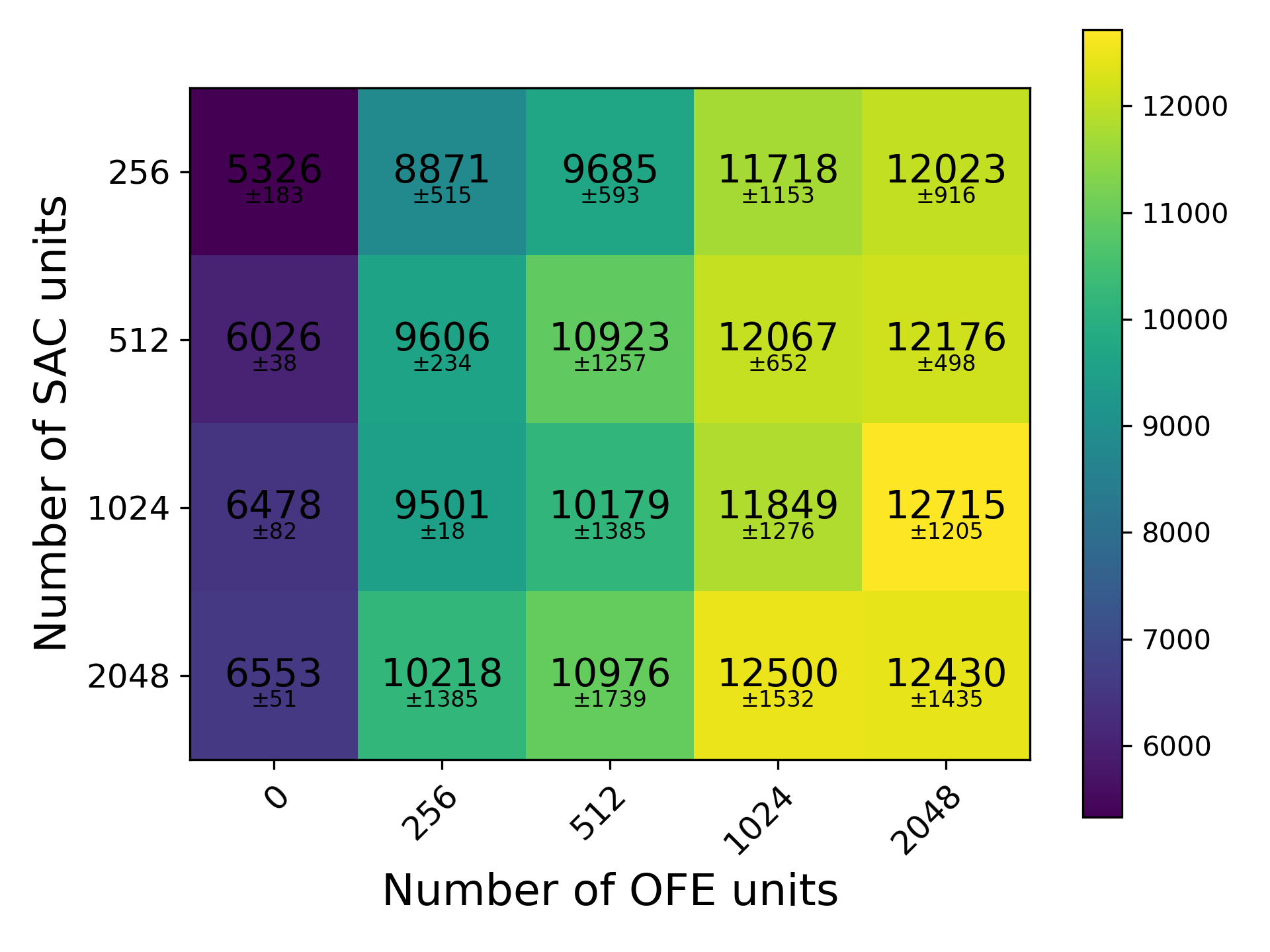}
    \vskip -0.2in
    \caption{Grid search results of average maximum return over different number of units between SAC and OFENet with ApeX-like distributed training. Compared to \fref{fig:ofe_sac_grid_search}, adding distributed RL enables monotonic improvement when we widen either SAC or OFENet.}
    \label{fig:apex_gridsearch}
\end{figure}

The results in \fref{fig:ofenet} shows separating representation learning from RL improves control performance and mitigates rank collapse of Q-networks regardless of network size.
Thus, we can conclude using bigger representations, which is learned using the auxiliary task (see \sref{sec:method_ofenet}), contributes to improve performance on downstream RL tasks.

To investigate more in-depth, we also conduct a grid search over different number of units for both SAC and OFENet in \fref{fig:ofe_sac_grid_search}. The baseline is SAC agent without OFENet (see leftmost column).
The results suggest that the performance does improve when compared against baseline agent (see horizontally), however, it saturates around the average return of $8000$.
In the following experiments, we employ distributed replay and expect we can attain higher performance.

\begin{figure*}[t]
	\begin{minipage}[]{0.32\linewidth}
		\centering
		\includegraphics[width=\columnwidth]{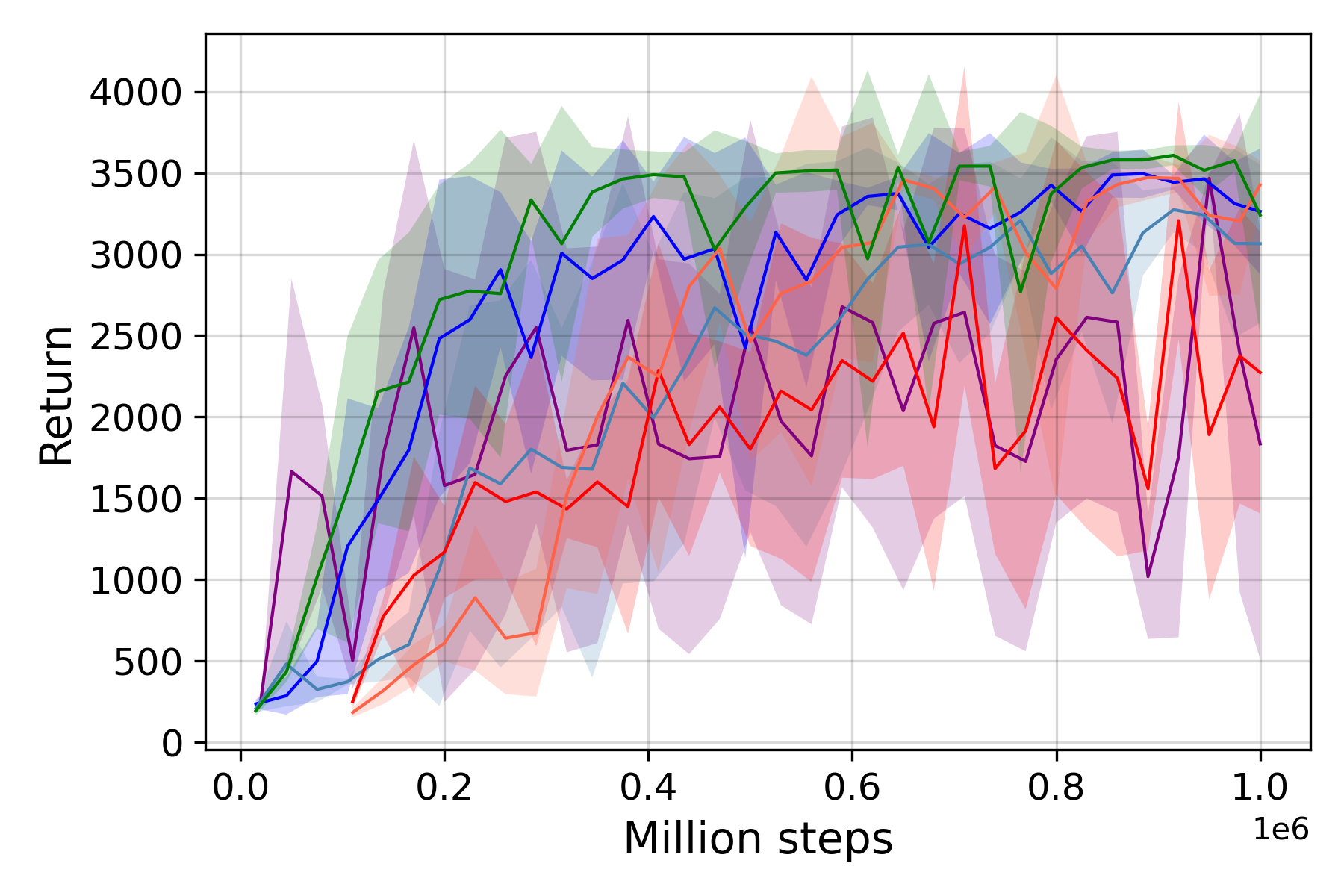}
		\subcaption{Hopper-v2}\label{fig:hopper}
	\end{minipage}
	\begin{minipage}[]{0.32\linewidth}
		\centering
		\includegraphics[width=\columnwidth]{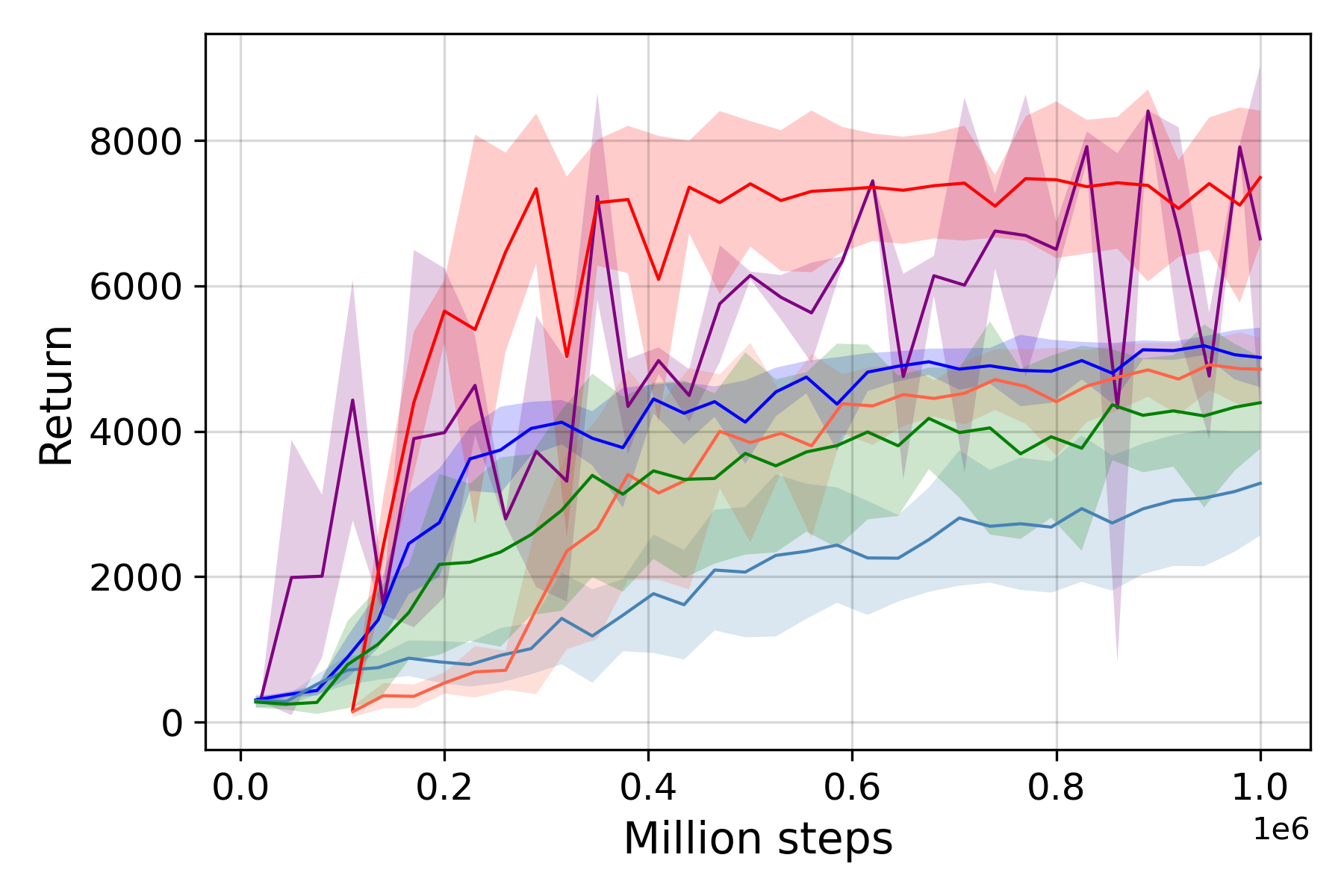}
		\subcaption{Walker2d-v2}\label{fig:walker2d}
	\end{minipage}
	\begin{minipage}[]{0.32\linewidth}
		\centering
		\includegraphics[width=\columnwidth]{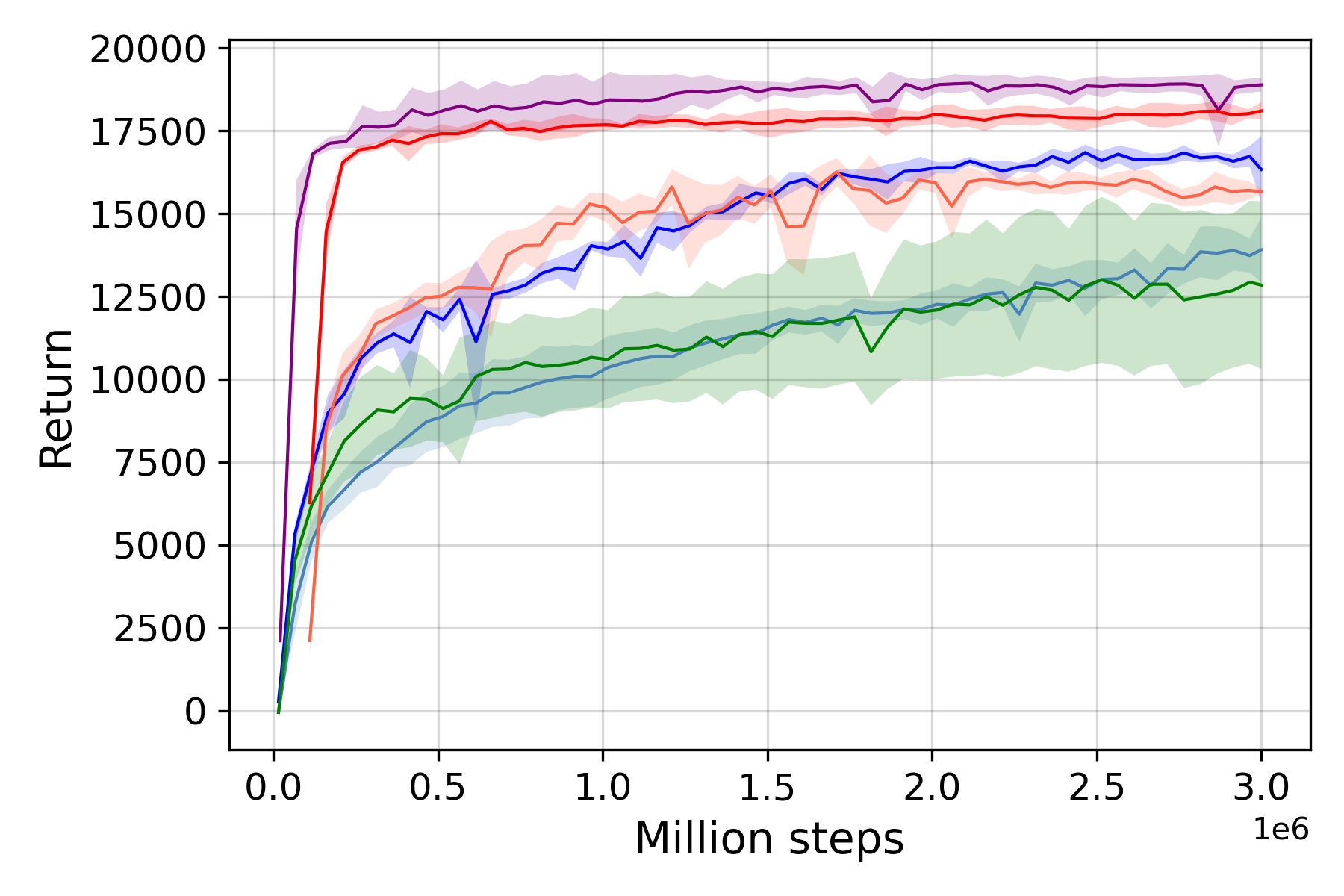}
		\subcaption{HalfCheetah-v2}\label{fig:halfcheetah}
	\end{minipage} \\
	\begin{minipage}[]{0.32\linewidth}
		\centering
		\includegraphics[width=\columnwidth]{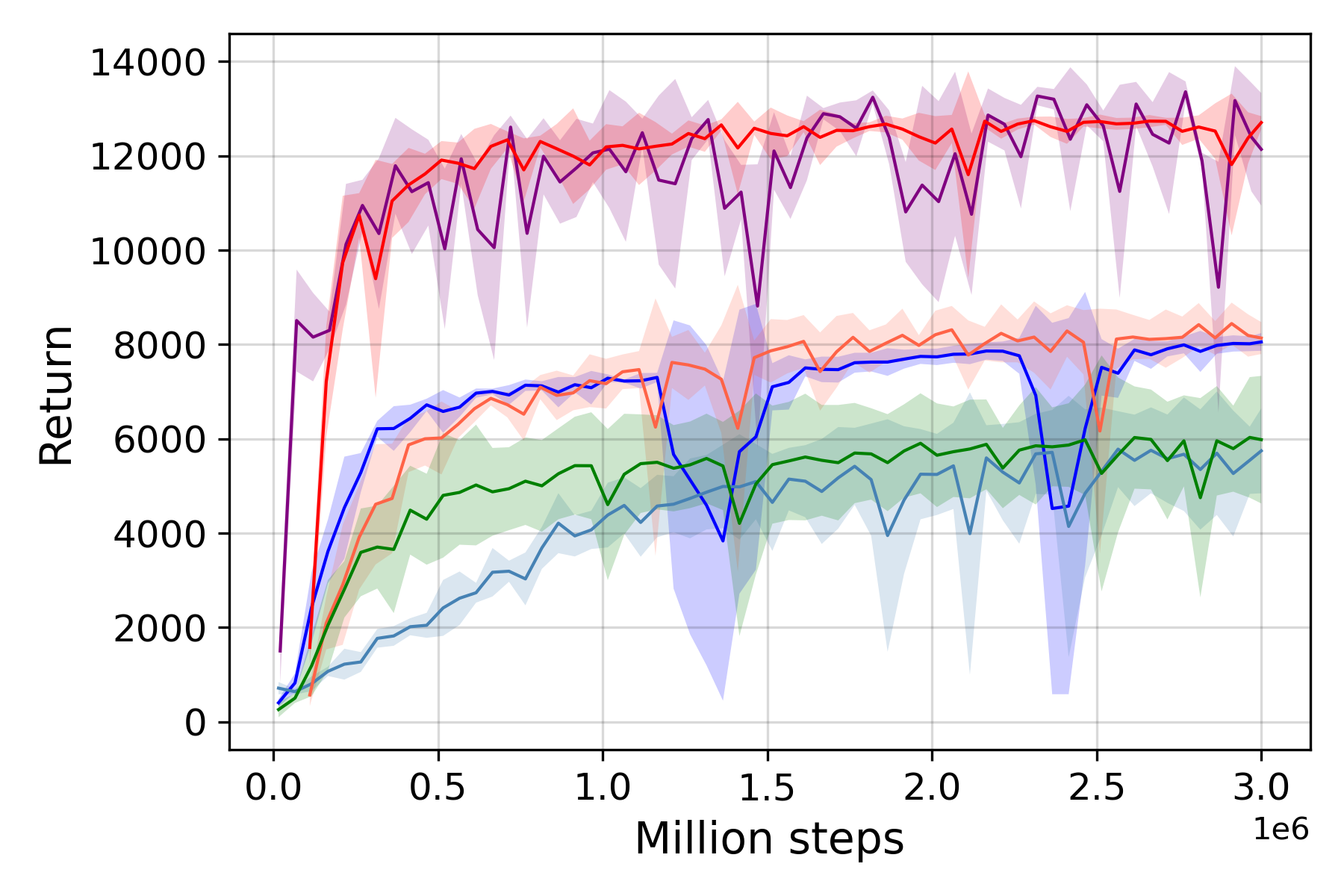}
		\subcaption{Ant-v2}\label{fig:Ant}
	\end{minipage}
	\begin{minipage}[]{0.32\linewidth}
		\centering
		\includegraphics[width=\columnwidth]{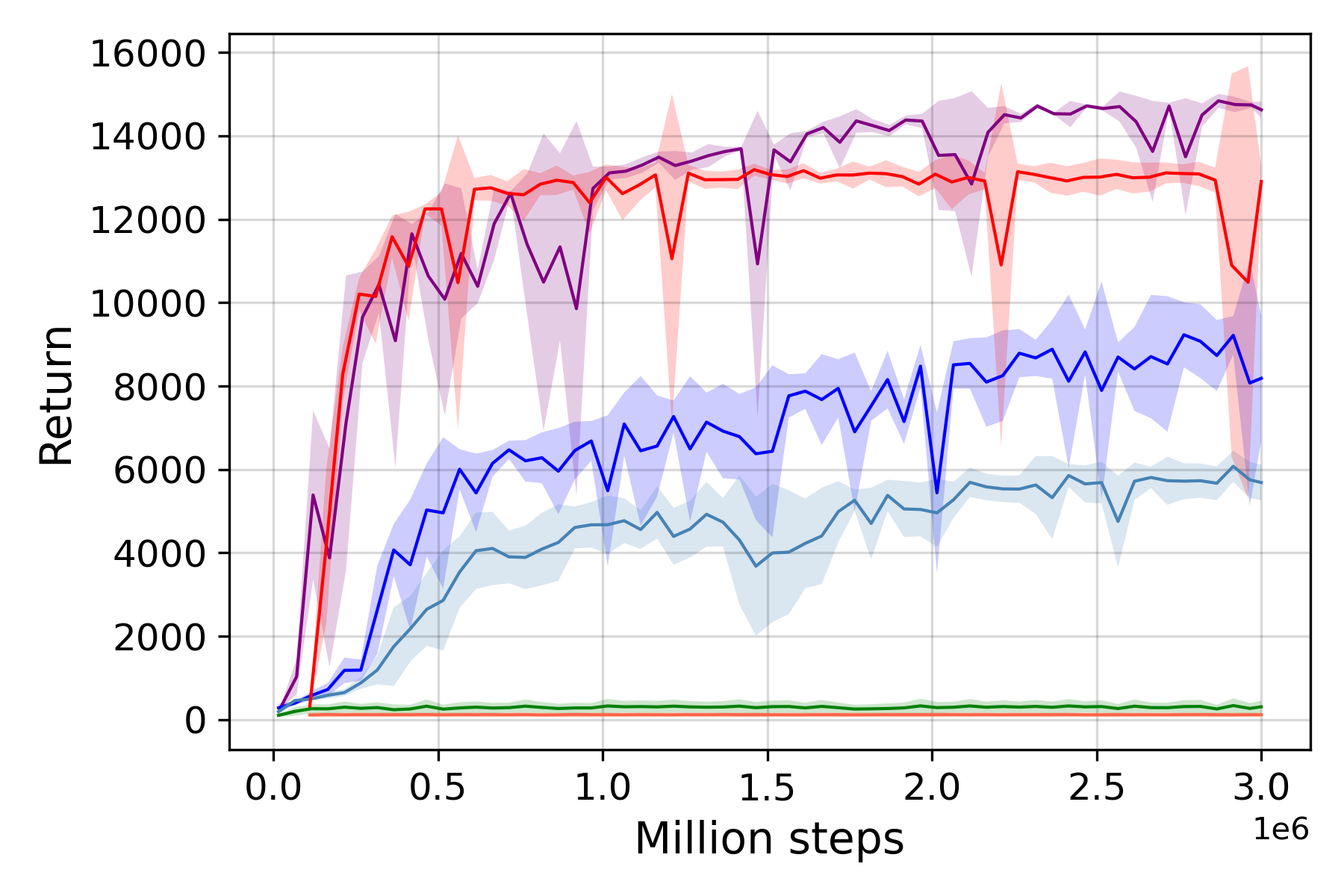}
		\subcaption{Humanoid-v2}\label{fig:humanoid}
	\end{minipage}
	\begin{minipage}[]{0.32\linewidth}
	\centering
	\includegraphics[width=2.8cm]{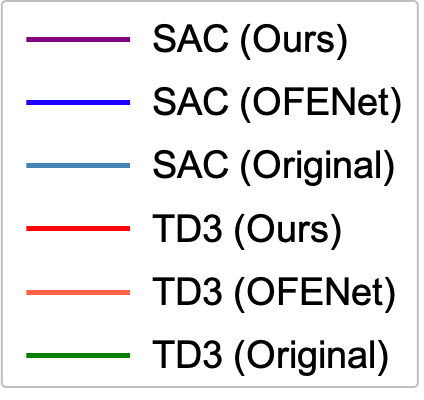}
    \end{minipage}	

	\caption{Training curves on five different MuJoCo tasks with two different RL algorithms (SAC and TD3).}
	\label{fig:all_env}
\end{figure*}
\begin{table*}[t]
    \caption{The highest average returns for each environment. The bold number indicates the best performance. Our method outperforms OFENet~\cite{ota2020can} and original algorithm in most environments.}
    \vskip 0.15in
    \begin{center}
    \begin{small}
    \begin{sc}
    \begin{tabular}{lrrrrrrr}
        \toprule
         & \multicolumn{3}{c}{SAC} & & \multicolumn{3}{c}{TD3} \\
                    \cmidrule{2-4} \cmidrule{6-8}
        Environment & \textcolor{Plum}{Ours} & \textcolor{blue}{OFENet} & \textcolor{TealBlue}{Original} & & 
                      \textcolor{red}{Ours} & \textcolor{RedOrange}{OFENet} & \textcolor{OliveGreen}{Original} \\
         \midrule
         Hopper-v2      & 3467.3        & {\bf 3511.6} & 3316.6 &
                        & 3206.7        & 3488.3       & {\bf 3613.0} \\
         Walker2d-v2    & {\bf 8802.4}  & 5237.0       & 3401.5 &
                        & {\bf 7645.8}  & 4915.1       & 4515.6  \\
         HalfCheetah-v2 & {\bf 19209.9} & 16964.1      & 14116.1 &
                        & {\bf 18147.5} & 16259.5      & 13319.9 \\
         Ant-v2         & {\bf 14021.0} & 8086.2       & 5953.1 &
                        & {\bf 12811.3} & 8472.4       & 6148.6 \\
         Humanoid-v2    & {\bf 14858.2} & 9560.5       & 6092.6 &
                        & {\bf 13282.0} & 120.6        & 340.5  \\
         \bottomrule
    \end{tabular}
    \end{sc}
    \end{small}
    \end{center}
    \vskip -0.1in
    \label{tab:all_env}
\end{table*}

\paragraph{Distributed RL}
Finally, we add distributed replay~\cite{horgan2018distributed} to further improve performance while using larger networks.
We use an implementation similar to~\cite{stooke2018accelerated}, which collects experiences using $N^\text{core}$ cores on which each core contains $N^\text{env}$ environments, specifically we used $N^\text{core}=2$ and $N^\text{env}=32$.

Similar to the previous experiments, we conduct a grid search over different number of units for SAC and OFENet with the distributed replay in \fref{fig:apex_gridsearch}, and also compare the training curves of three different network size \emph{S}, \emph{M}, and \emph{L} in~ \Apref{appendix:additional_results}.
Comparing \fref{fig:apex_gridsearch} and \fref{fig:ofe_sac_grid_search}, we can clearly see the distributed training enables further performance gain on all network size.
Furthermore, we can observe monotonic improvement when we increase the number of units for both SAC and OFENet.
Thus, we verified combining distributed replay contributes further performance gain while training larger networks.



\paragraph{How about generalization to different RL algorithms and environments?}
To quantitatively measure the effectiveness of our method across different RL algorithms and tasks, we evaluate two popular optimization algorithms, namely SAC and TD3~\cite{fujimoto2018addressing}, on five different locomotion tasks in MuJoCo~\cite{todorov2012mujoco}.
We denote our method as \emph{Ours}, which uses the largest network of $N^\text{units}=2048$ among the previous experiments for both the OFENet and the RL algorithms.
We compare the proposed method against two baselines: the original RL algorithm denoted by \emph{Original}.
Furthermore, we also compare OFENet, which can achieve the current state-of-the-art performance on these tasks to the best of our knowledge.

We plot the training curves in \fref{fig:all_env} and list the highest average return in \Tref{tab:all_env}. In the figure and the table, our method, \emph{SAC (Ours)} and \emph{TD3 (Ours)} achieves the best performance on almost all environments. Furthermore, we can see that our proposed method can work with both RL algorithms, and thus is agnostic to the choice of the trainign algorithm.
In particular, our method notably achieves much higher episode return in Ant-v2 and Humanoid-v2, which are harder environments with larger state/action space and more training examples.
Interestingly, the proposed method does not achieve reasonable solutions in Hopper-v2, which has the smallest dimensionality among five environments. We consider that the performance in smaller dimension problem saturates early and even additional methods are unable to provide any significant performance gain.



\subsection{Ablation study}\label{sec:exp_ablation}
\newcommand{\full}{{\it Full }}
\newcommand{\noapex}{{\it w/o Ape-X }}
\newcommand{\noofe}{{\it w/o OFENet }}
\newcommand{\nolarge}{{\it w/o Larger NN }}
\newcommand{\nodensenet}{{\it w/o DenseNet }}
\newcommand{\sac}{{\it sac }}
Since our method integrates several different ideas into a single agent, we conduct additional experiments to understand what components contribute to the performance gain.
We highlight that our method consists of three elements: feature representation learning using OFENet, DenseNet architecture, and distributed training.
In addition to this, we compare the results without increasing the network size to reinforce that larger network does improve performance.
\Fref{fig:ablation_study} shows the ablation study over SAC with Ant-v2 environment.
\full is our method which combines all three elements we proposed, and uses large networks ($N^\text{unit}=2048,N^\text{layer}=2$) for the SAC agent. \sac is the original SAC implementation.

\noapex removes Ape-X-like distributed training setting.
As distributed RL enables collection of more experiences close to the current policy, we consider that the significant performance gain can be explained by learning from more on-policy data, which was also empirically shown by~\cite{fedus2020revisiting}.
Also, we believe that receiving more novel experiences helps the agent generalize to state-action space. In other words, more novel experience reduces overfitting to limited trajectories, which becomes more important in harder environments which has larger state/action space, and larger neural networks.


\noofe removes OFENet and trains the whole architecture by using only a scalar reward signal.
The much lower return shows that learning the large networks from just the scalar reinforcement signal is difficult, and training the bottom networks (close to the input layer), i.e., obtaining informative features by using an auxiliary task enables better learning of control policy.

\nolarge reduces the number of units from $N^\text{unit}=2048$ to $256$ for both OFENet and SAC.
This also significantly drops the performance, and thus we can conclude that using larger networks is essential to achieve high performance.

Finally, \nodensenet replaces MLP-DenseNet defined in \sref{sec:method_architecture} with standard MLP architecture.
The result shows that strengthening feature propagation does contribute to improve performance.

\begin{figure}
    \centering
    \includegraphics[width=0.8\columnwidth]{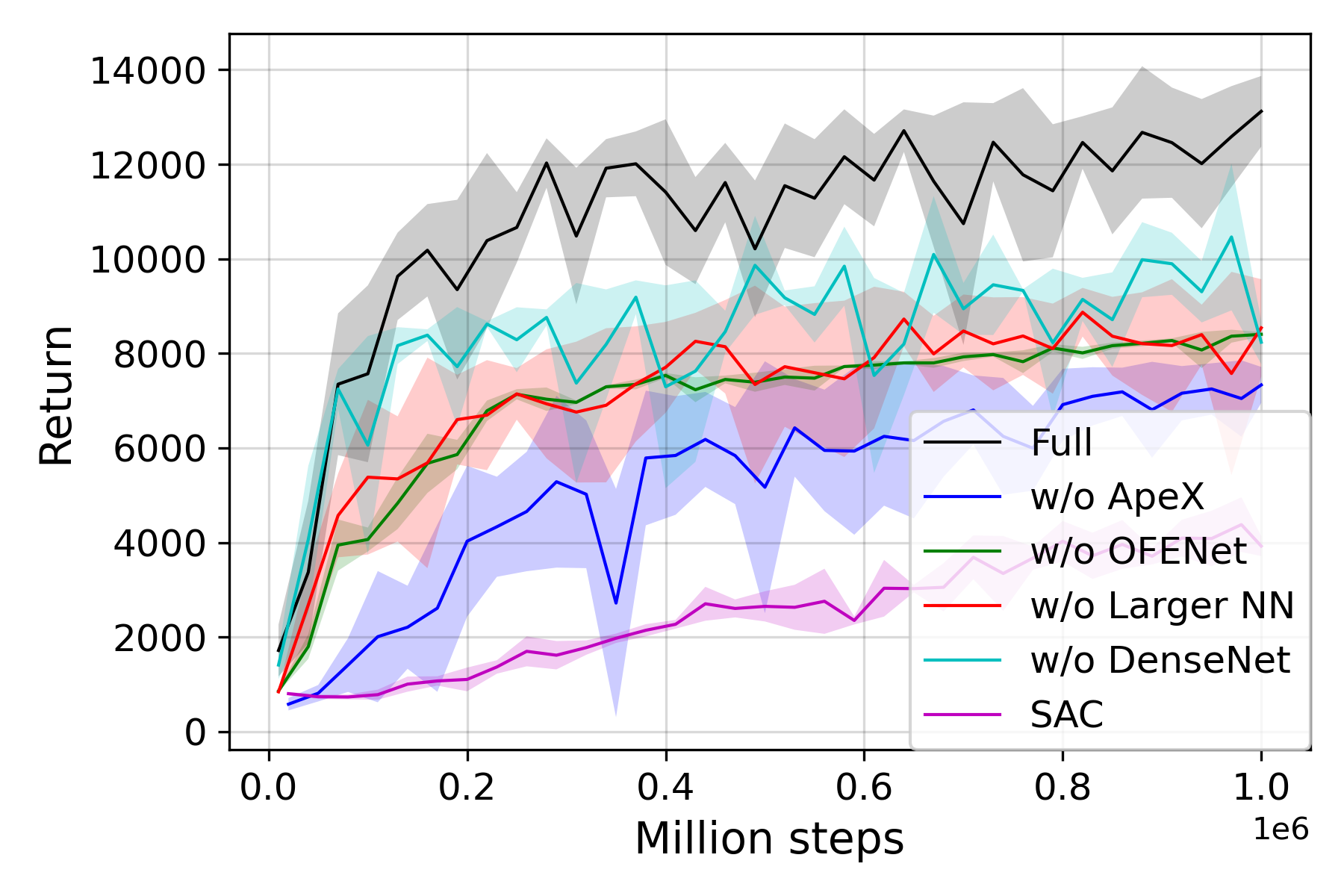}
    \vskip -0.1in
    \caption{Training curves of the derived methods of SAC on Ant-v2. This shows that each element does contribute to performance gain, and our combination of DenseNet architecture, distributed training, and decoupled feature representation (shown as \full) allows us to train larger networks that performs significantly better compared against the baseline SAC algorithm (shown as \sac).}
    \label{fig:ablation_study}
\end{figure}

    \section{Conclusion}\label{sec:conclusions}
Deep Learning has catalyzed huge breakthroughs in the fields of computer vision and natural language processing making use of massive neural networks that can be trained with huge amounts of data. While these domains have hugely benefitted from the use of larger networks, the RL community has not witnessed similar trend in use of larger networks for training high performance agents. This is mostly due to instability that occurs when using larger networks for training RL agents. In this paper, we studied the problem of using larger network for training RL agents. To achieve this, we proposed a novel method for training larger networks for deep RL agents while reflecting on some of the important design choices one has to make when using such networks. In particular, the proposed method consists of three elements.
First, we decouple representation learning from RL using an auxiliary loss of predicting the next state. This allows to obtain more informative features to be used to learn control policies with richer information compared to learning entire networks from a scalar reward signal.
The learned representation is then propagated to the DenseNet architecture that consists of very wide networks.
Finally, a distributed training framework provides huge amounts of on-policy data whose distribution is much closer to the current policy, and thus enables to mitigate overfitting problem and enhance generalization to novel scenarios.
Our experiments demonstrate that this novel combination achieves significantly higher performance compared against the current state-of-the-art algorithms across different off-policy RL algorithms and different continuous control tasks.

In the future, we would like to study the application to high-dimensional inputs (e.g., images). We also would like to
investigate how we can make use of the proposed method for other off-policy methods to make our method agnostic to the choice of underlying algorithm.



    \clearpage
    \bibliography{cite}
    \bibliographystyle{icml2020}

    \clearpage
    \onecolumn
    \appendix
\section{Experimenal Details} \label{appendix:exp_details}
This section describes more details of our experiments.

\subsection{Implementation}
\paragraph{OFENet}
To implement OFENet, we referred the official codebase provided by~\citet{ota2020can}, which is available at their website\footnote{Codes used for implementing MLP-ResNet and MLP-DenseNet can be found at \url{https://www.merl.com/research/license/OFENet}}.
We also employed target networks~\cite{mnih2015human} to stabilize the training of OFENet, since the distribution of experiences stored in the shared replay buffer can change more dynamically by utilizing the Ape-X-like distributed training setting as described in~\sref{sec:method_dist_sample}.
The target networks are updated on each training step by having them slowly track the learned networks: $\theta^{\prime} \leftarrow \tau \theta+(1-\tau)\theta^{\prime}$, where we assume $\theta$ to be the network parameters of the current OFENet, and $\theta^{\prime}$ is the parameters of the target network. We use the target smoothing coefficient $\tau=0.005$, which is the same with the one used to update target value networks in SAC~\cite{haarnoja2018soft}, in other words, we do not tune this parameter.

\paragraph{RL agents}
Our implementation of the RL agents are also based on the public codebase used in~\cite{ota2020can}.
As for Batch Normalization~\cite{ioffe2015batch}, which we use for OFENet and \emph{TD3 (Ours)} in~\fref{fig:all_env}, we use its training mode in updating the network, and test mode in collecting experiences as done in~\cite{liu2021regularization}.
We also used Huber loss to stabilize the training of RL agents for the same reason that we employ the target network for training OFENet described in the previous paragraph.

\begin{figure}[h]
	\begin{center}
		\includegraphics[clip,width=0.6\linewidth]{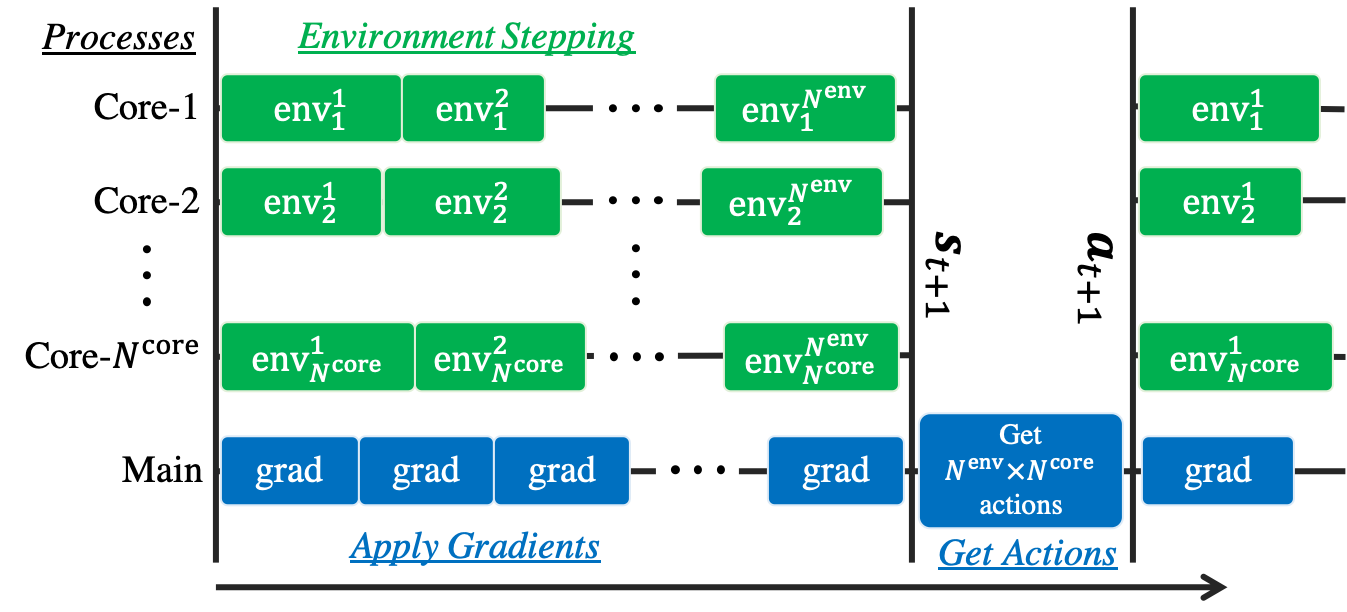}
		\caption{Schematic of asynchronous training. We use $N^\text{core}=2$ cores for collecting experiences, where each core has $N^\text{env}=32$ environments. Since the network parameters are shared and the training and collecting transitions are decoupled, the collected experiences result in more on-policy data compared against the standard off-policy training, where the agent collects one transition while it applies one gradient step.}
		\label{fig:asynchronous_trianing}
	\end{center}
\end{figure}

\paragraph{Distributed training}
The distributed training setting we used is similar to~\cite{stooke2018accelerated}, which collects experiences using $N^\text{core}$ cores on which each core contains $N^\text{env}$ environments. Specifically, we used $N^\text{core}=2$ and $N^\text{env}=32$.
\Fref{fig:asynchronous_trianing} shows the schematic of the distributed training. Since the actions are computed by the latest parameters, the collected experiences result in more on-policy data.

\clearpage
\subsection{Network Architectures} \label{appendix:architectures}
We highlight that our proposed architecture consists of three elements: 1) Decoupling representation learning from RL using OFENet, 2) DenseNet architecture with large NNs to effectively propagate the features obtained using OFENet, and 3) distributed training to obtain more on-policy experiences that can mitigate overfitting problems and improve performance (see~\fref{fig:architecture} as well).

As described in~\sref{sec:method_architecture}, the DenseNet architecture consists of a composite function of fully-connected layer, Batch Normalization, and an activation function.
We choose the activation function to be Swish~\cite{ramachandran2017searching} for MLP-ResNet and MLP-DenseNet, because it showed the smallest value of the auxiliary loss, i.e. attains the best accuracy of function approximation for predicting the next state on all environments as shown in~\cite{ota2020can}. (We compared the performance of RL with different activation functions in~\fref{fig:compare_activation})

For OFENet, we designed the architecture to increase the feature dimensionality to $2048$ from original inputs by using 8-layers DenseNet architecture as proposed in~\cite{ota2020can}.
For SAC agent, we designed the architecture to concatenate $2048$ features at each layer.
\Tref{tab:param_sac_ofe} shows the number of parameters for \emph{SAC (Ours)} and \emph{SAC (Original)} as reference on Ant-v2 environment as an example, which has 111 and 8 dimensionality for state and action space.
As you can see from the table, we increase the parameters of the network to be 100 times more than the original SAC implementation.

\begin{table}[h]
    \caption{The parameter size of \emph{SAC (Ours)} and \emph{SAC (Original)} used in experiments on Ant-v2 environment in \sref{sec:architecture_comparison} and \sref{sec:exp_ablation}.}
    \vskip 0.15in
    \begin{center}
    \begin{small}
    \begin{sc}
    \begin{tabular}{llrrrrrrr} 
        \toprule
         & & \multicolumn{3}{c}{SAC (Ours)} & & \multicolumn{3}{c}{SAC (Original)} \\
         \cmidrule{3-5} \cmidrule{7-9}
         & & Input & Output & \multirow{2}{*}{Parameters} & & Input & Output & \multirow{2}{*}{Parameters} \\
         & & units & units  &                             & & units & units  & \\
         \midrule
         \multirow{9}{*}{OFENet: $z_s$}
         & $1$st layer & $111$ & $256$ & 29,696 \\
         & $2$nd layer & 367 & $256$ & 95,232 \\
         & $3$rd layer & 623 & $256$ & 160,768 \\
         & $4$th layer & 879 & $256$ & 226,304 \\
         & $5$th layer & 1,135 & $256$ & 291,840 \\
         & $6$th layer & 1,391 & $256$ & 357,376 \\
         & $7$th layer & 1,647 & $256$ & 422,912 \\
         & $8$th layer & 1,903 & $256$ & 488,448 \\
         & Total & & & $\mathbf{2,072,576}$\\
         \midrule
         \multirow{4}{*}{SAC}
         & $1$st layer & 2,159 & 2,048 & 4,423,680 & & 111 & 256 & 28,672\\
         & $2$nd layer & 4,207 & 2,048 & 8,617,984 & & 256 & 256 & 65,792\\
         & Output layer & 6,255 & 8 & 50,048     & & 256 & 8   & 2,056\\
         & Total & & & $\mathbf{13,091,712}$ & & & & $\mathbf{96,520}$\\
         \midrule
          & Total & & & $\mathbf{10,709,000}$ & & & & $\mathbf{96,520}$ \\
         \bottomrule
    \end{tabular}
    \end{sc}
    \end{small}
    \end{center}
    \vskip -0.1in
    \label{tab:param_sac_ofe}
\end{table}

\clearpage
\subsection{Visualizing loss surface of Q-function networks}\label{appendix:deeper_wider_surface}
This section provides the details of how we estimate the loss surface shown in~\fref{fig:sac_deeper_curvature} and~\fref{fig:sac_wider_curvature}.

\citet{li2018visualizing} proposed a method to visualize the loss function curvature by introducing \emph{filter normalization} method. The authors empirically demonstrated the non-convexity of the loss functions can be problematic, and the sharpness of the loss surface correlates well with test error and generalization error.
In light of this, we also visualize the loss surface of the networks to figure out why the deeper network could not lead to better performance while the wider networks result in high performance policies (\fref{fig:grid_search_ant_sac_woofe_raw}).

To visualize the loss surface of our Q-networks, we use the authors' implementation\footnote{Code used for these plots can be found at: \url{https://github.com/tomgoldstein/loss-landscape}} with the loss of:
\begin{equation}
    J_{Q}(\theta)=\mathbb{E}_{\left(\mathbf{s}_{t}, \mathbf{a}_{t}\right) \sim \mathcal{D}}\left[\frac{1}{2}\left(Q_{\theta}\left(\mathbf{s}_{t}, \mathbf{a}_{t}\right)-\hat{Q}\left(\mathbf{s}_{t}, \mathbf{a}_{t}\right)\right)^{2}\right],
\end{equation}
with 
\begin{equation}
    \hat{Q}\left(\mathbf{s}_{t}, \mathbf{a}_{t}\right)=r\left(\mathbf{s}_{t}, \mathbf{a}_{t}\right)+\gamma \mathbb{E}_{\mathbf{s}_{t+1} \sim p}\left[V_{\bar{\psi}}\left(\mathbf{s}_{t+1}\right)\right],
\end{equation}
in which we exactly follow the notations used by SAC paper~\cite{haarnoja2018soft}.
To compute this objective $J_{Q}(\theta)$, we collect all transitions used in the training of the deeper and wider networks, and compute the target values of $\hat{Q}\left(\mathbf{s}_{t}, \mathbf{a}_{t}\right)$ after finishing training and stored the tuples of $\left( s_t, a_t, \hat{Q}\left(\mathbf{s}_{t}, \mathbf{a}_{t}\right) \right)$ for all transitions in the training.
Then, we use the authors' implementation to visualize the loss with the stored transitions and trained weights of the  Q-network.
Please refer to \cite{li2018visualizing} for more details.



\subsection{Hyperparameters}
\paragraph{OFENet}
All OFENet networks we used for our experiments consist of 8-layers DenseNet architectures with Swish activation~\cite{ramachandran2017searching} as used in~\cite{ota2020can}.
The output dimensionality of OFENet is defined in each experiment. \Tref{tab:param_sac_ofe} in the previous section shows the detailed output units in each layer as an example.

\paragraph{RL algorithms}
The hyperparameters of the RL algorithms are also the same with their original papers, except that the TD3 uses the batch size $256$ instead of $100$ as done in~\cite{ota2020can}.
Also for fair comparison to~\cite{ota2020can}, we used a random policy to store transitions to replay buffer before training RL agents for 10K time steps for SAC, and for 100K steps for TD3.

\paragraph{Effective rank}
We employ the \emph{effective rank}, which is recently proposed in~\cite{kumar2021implicit}, as a metric to evaluate the effectiveness of our architecture as described in~\sref{sec:methods}.
Following the notations given by~\citet{kumar2021implicit}, the effective rank can be computed as $\operatorname{srank}_{\delta}(\Phi)=\min \left\{k: \frac{\sum_{i=1}^{k} \sigma_{i}(\Phi)}{\sum_{i=1}^{d} \sigma_{i}(\Phi)} \geq 1-\delta\right\}$, where $\left\{\sigma_{i}(\Phi)\right\}$ are the singular values of feature matrix $\Phi$, which is the features of the penultimate layer of the Q-networks.
We used $\operatorname{srank}_{\delta}(\Phi)=0.01$ to calculate the number of effective ranks in the experiments, as in~\cite{kumar2021implicit}.

\clearpage
\section{Additional Results} \label{appendix:additional_results}
\subsection{Training Curves}

\paragraph{Distributed RL}
In order to evaluate the performance of distributed RL, we compare the performance of our method w/ and w/o Ape-X-like distributed training over different three network size: $N^\text{units} \in \left \{ 256, 1024, 2048 \right\}$, which we respectively denote \emph{S}, \emph{M}, and \emph{L}.
We denote the baseline by \emph{w/o Ape-X}, which are the same with \emph{w/ OFENet} in \fref{fig:ofenet}. It is noted again that the horizontal axis indicates the number of steps we applied gradients, not environmental steps as we use distributed actors that interact with environments in parallel (see~\fref{fig:architecture}).

\begin{figure}[h]
    \centering
    \includegraphics[width=0.5\columnwidth]{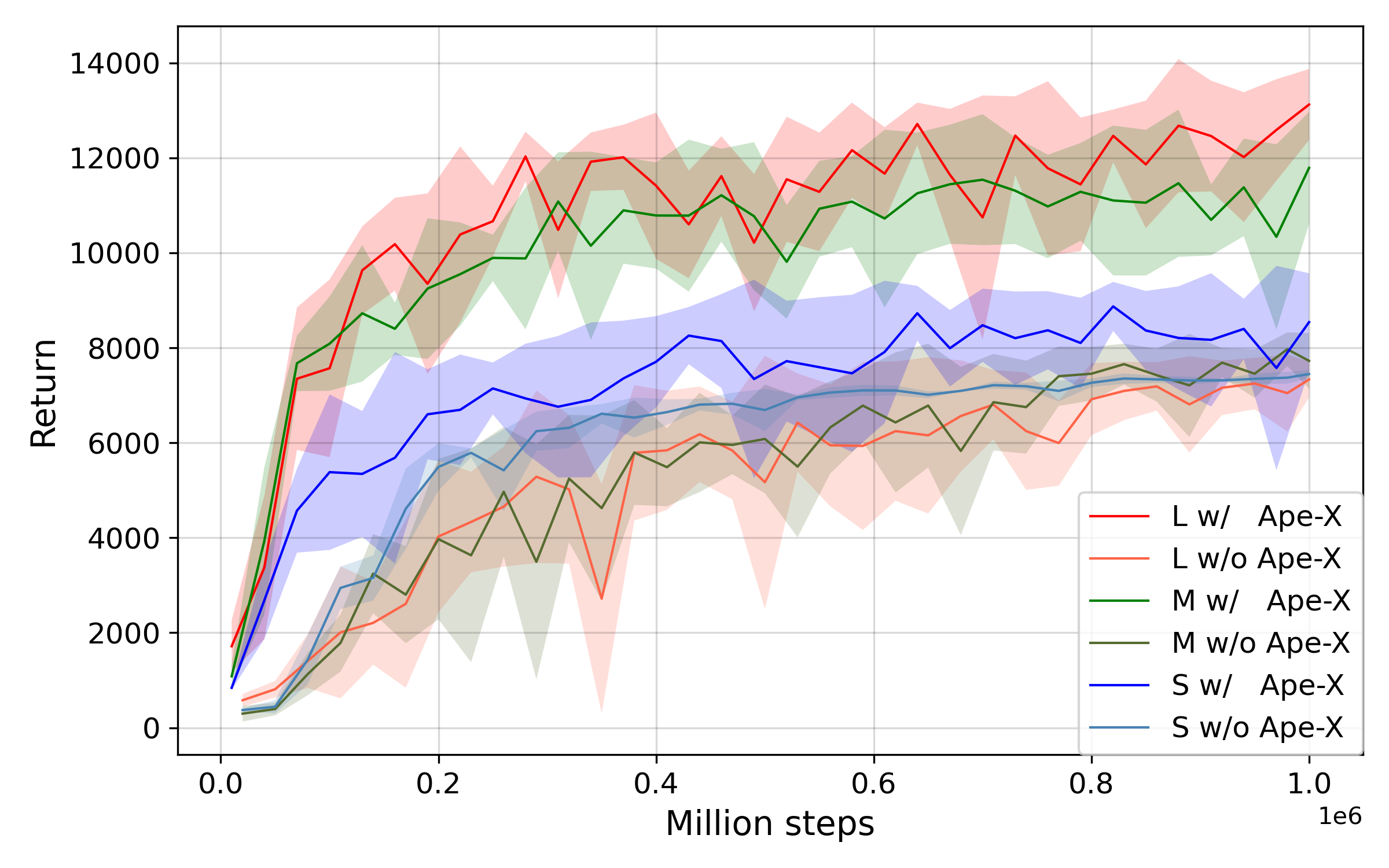}
    \caption{Comparison of w/ and w/o using Ape-X architecture.}
    \label{fig:apex_training_curves}
\end{figure}

\Fref{fig:apex_training_curves} shows that using Ape-X enables to improve performance on all network size, and the larger networks tend to further improve the performance.

\begin{figure*}[h]
	\begin{minipage}[]{0.32\linewidth}
		\centering
		\includegraphics[width=\columnwidth]{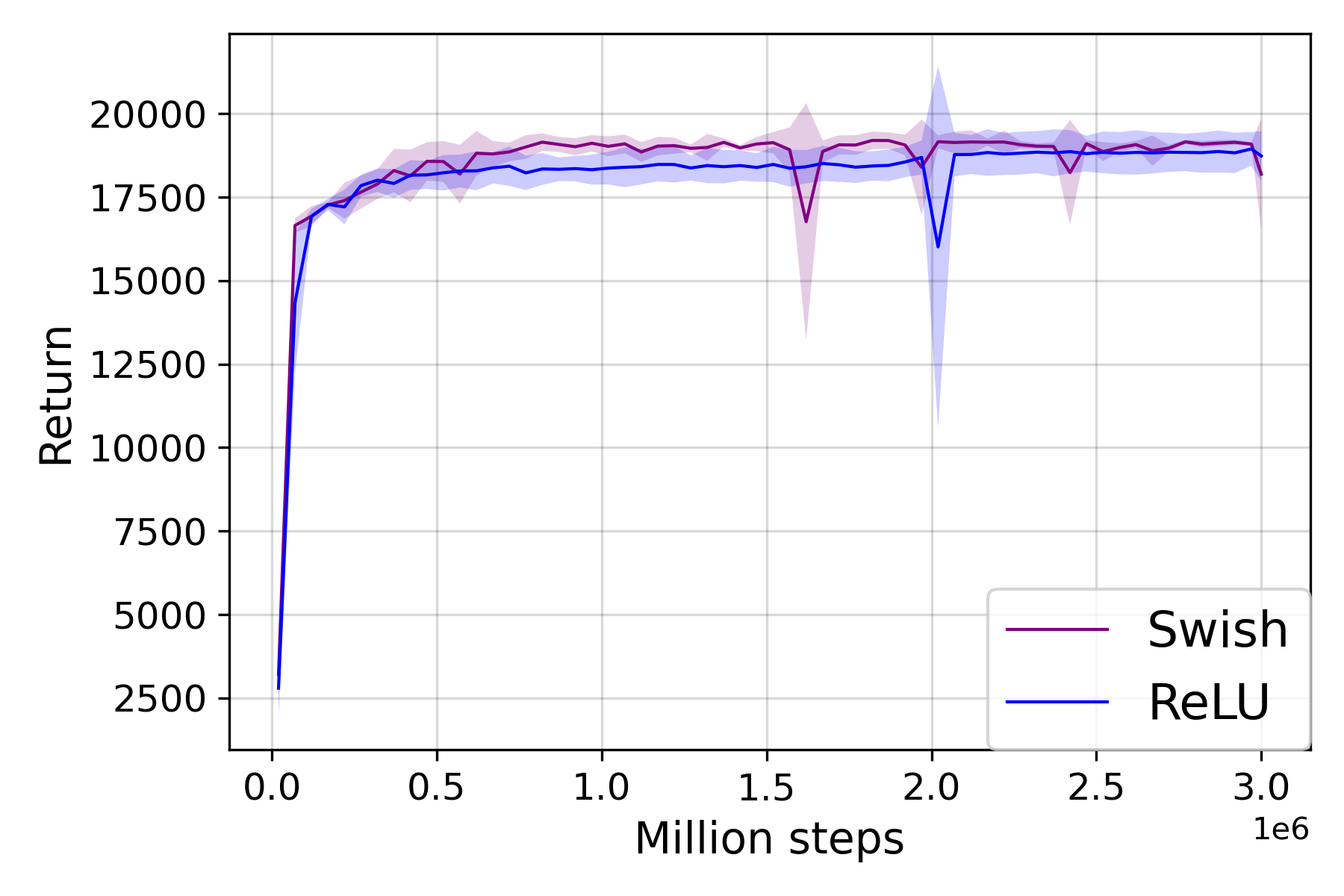}
		\subcaption{HalfCheetah-v2}
	\end{minipage}
	\begin{minipage}[]{0.32\linewidth}
		\centering
		\includegraphics[width=\columnwidth]{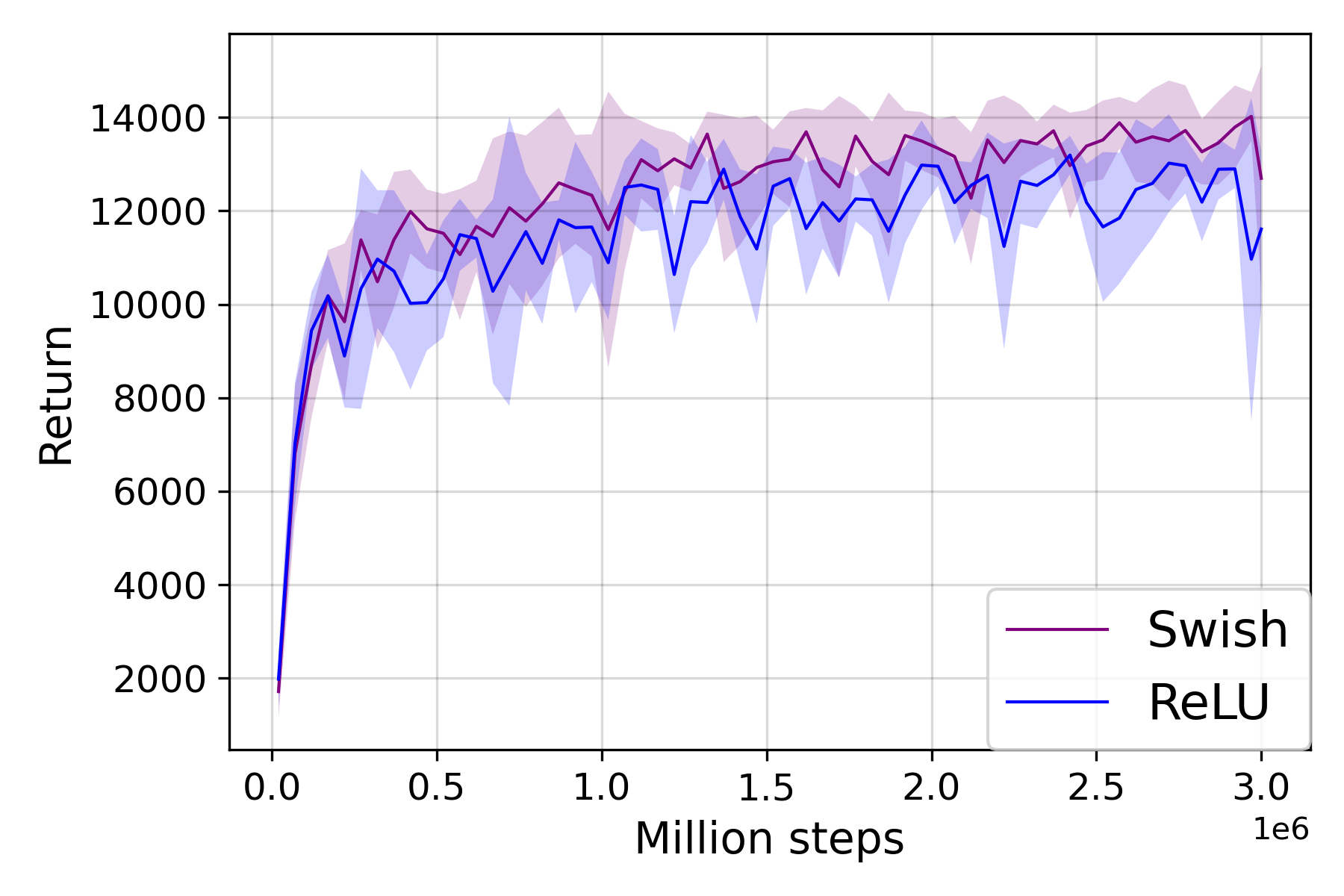}
		\subcaption{Ant-v2}
	\end{minipage}
	\begin{minipage}[]{0.32\linewidth}
		\centering
		\includegraphics[width=\columnwidth]{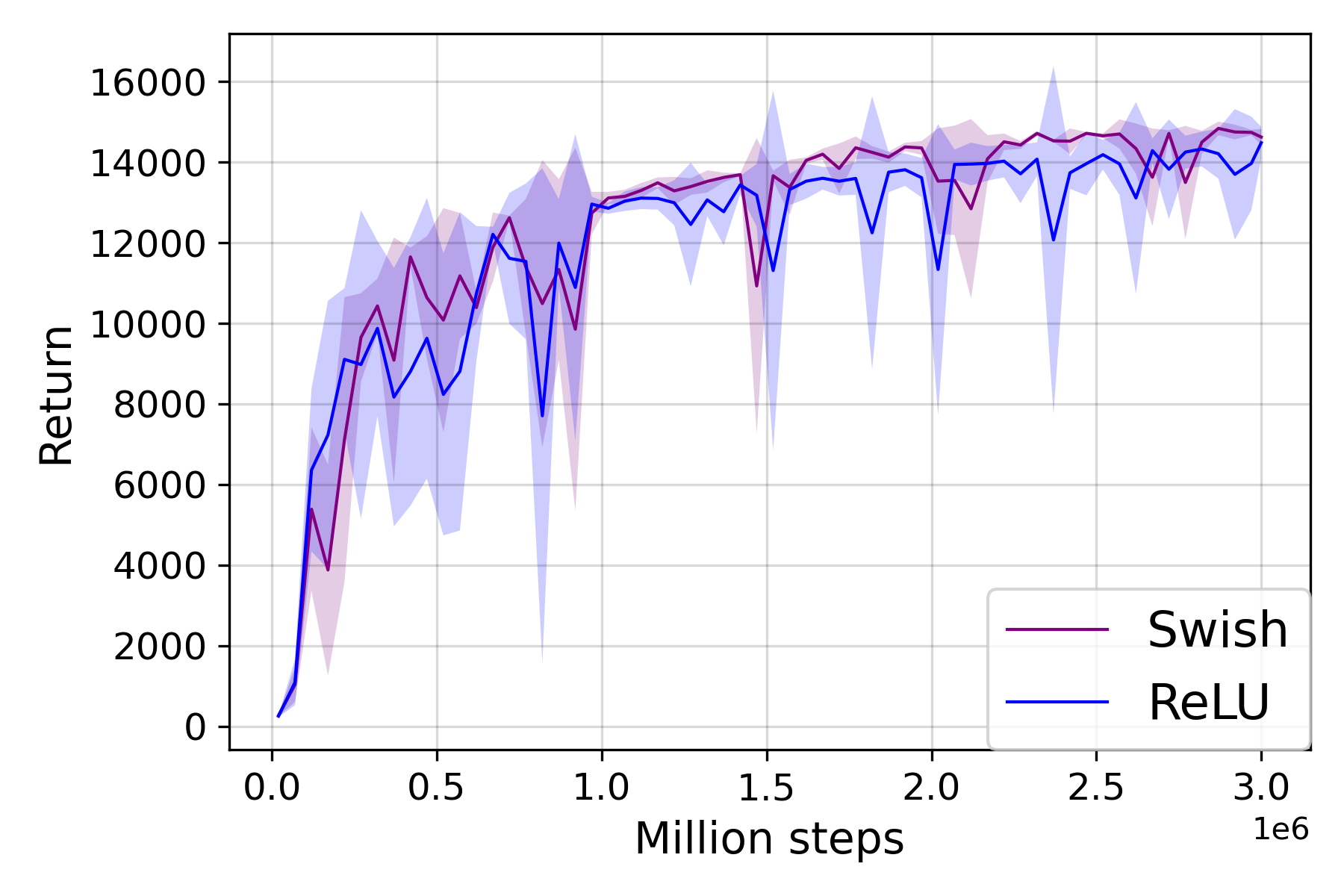}
		\subcaption{Humanoid-v2}
	\end{minipage}

	\caption{Comparison of activation function of DenseNet architecture between Swish and ReLU.}
	\label{fig:compare_activation}
\end{figure*}

\paragraph{Activation}
As described in \Apref{appendix:architectures}, we used Swish activation for RL agents (policy and value function networks) as it is shown to improve accuracy of function approximation in~\cite{ota2020can}.
To evaluate the effect of different activation functions, we plot the results of \emph{SAC (Ours)} with different activation functions (ReLU and Swish) for policy and value function networks used in SAC in \fref{fig:compare_activation}.
The results show we have slight performance gain by replacing ReLU with Swish.
More comprehensive empirical studies can be found in~\cite{andrychowicz2020matters,henderson2018deep}.

\clearpage
\subsection{Loss surface}
\begin{figure*}[h]
	\begin{minipage}[]{0.48\linewidth}
		\centering
		\includegraphics[width=0.8\columnwidth]{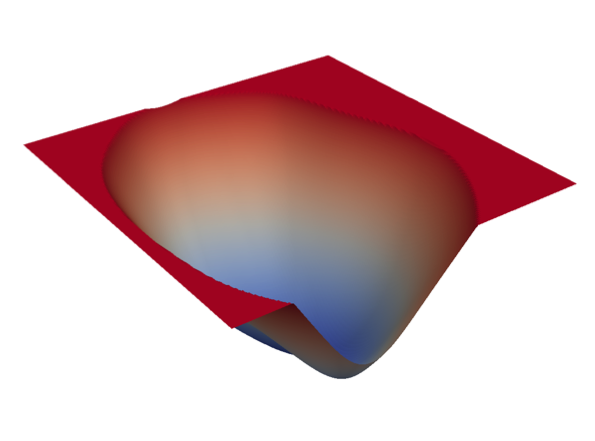}
		\subcaption{Wider network ($N^\text{layer}=2, N^\text{unit}=2048$)}\label{fig:loss_surface_wider_halfcheetah}
	\end{minipage}
	\begin{minipage}[]{0.48\linewidth}
		\centering
		\includegraphics[width=0.8\columnwidth]{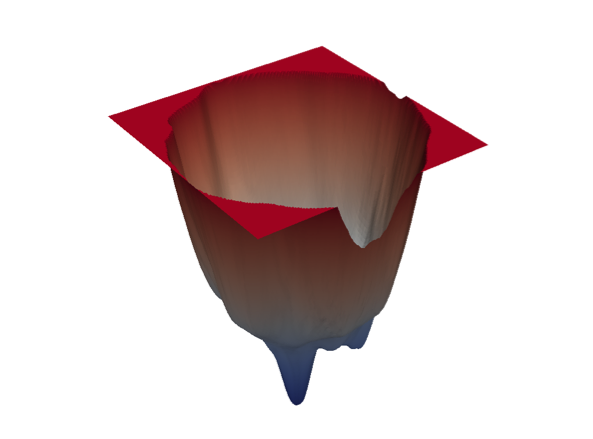}
		\subcaption{Deeper network ($N^\text{layer}=16, N^\text{unit}=256$)}\label{fig:loss_surface_deeper_halfcheetah}
	\end{minipage}
	\caption{Loss landscapes of models trained on HalfCheetah-v2 with one million steps, visualized using the technique in~\citet{li2018visualizing} and settings described in~\Apref{appendix:deeper_wider_surface}. }
	\label{fig:loss_surface_halfcheetah}
\end{figure*}

In addition to showing the loss surface of the deeper and wider networks that are trained on Ant-v2 environment in~\fref{fig:sac_deeper_curvature} and~\fref{fig:sac_wider_curvature}, we also visualize the loss landscapes of Q-networks trained on HalfCheeta-v2 environment in~\fref{fig:loss_surface_halfcheetah}.
Together with the results shown in~\fref{fig:sac_deeper_curvature} and~\fref{fig:sac_wider_curvature}, we can see the wider networks tend to converge to a \emph{flatter} minimum while the deeper networks have \emph{sharper} minimums.

    \include{appendix}
\end{document}